\documentclass[runningheads]{llncs}

\usepackage{eccv}

\usepackage{eccvabbrv}

\usepackage{graphicx}
\usepackage{booktabs}

\usepackage[accsupp]{axessibility}  
\usepackage[table]{xcolor}
\usepackage{tabularx}
\usepackage{multirow}    
\usepackage{makecell}    
\usepackage{xcolor}      
\usepackage{amssymb}     
\usepackage{graphicx}    
\usepackage{booktabs}
\usepackage{multirow}
\usepackage{colortbl}
\usepackage{booktabs}
\usepackage{multirow}
\usepackage[table]{xcolor}
\usepackage{array}  

\usepackage{hyperref}

\usepackage{orcidlink}
\usepackage{booktabs}
\usepackage{siunitx}
\usepackage{cellspace}
\usepackage[table]{xcolor}
\usepackage{pifont}
\usepackage{booktabs}
\usepackage{multirow}
\usepackage{pifont}
\usepackage[normalem]{ulem} 
\newcommand{\cmark}{\ding{51}}
\newcommand{\xmark}{\ding{55}}

\definecolor{sourcegray}{gray}{0.92}
\definecolor{oursgray}{gray}{0.85}

\begin{document}

\title{DAP: Doppler-aware Point Network for Heterogeneous mmWave Action Recognition} 

\titlerunning{DAP}


\author{
Jiaying Lin\inst{1}\orcidlink{0009-0009-1511-4151}\and
Shiman Wu\inst{2}\orcidlink{0009-0007-1256-8882}\and
Jinfu Liu\inst{3}\orcidlink{0009-0001-0994-9146}\and
Can Wang\inst{4,5} \and 
Mengyuan Liu\inst{1}$^\dagger$\orcidlink{0000-0002-6332-8316}
}

\authorrunning{J.~Lin et al.}

\institute{
$^{1}$State Key Laboratory of General Artificial Intelligence, Peking University, Shenzhen Graduate School, Shenzhen, China\\
$^{2}$Huazhong University of Science and Technology, Wuhan, China\\
$^{3}$DJI Technology Co., Ltd., Shenzhen, China\\
$^{4}$China Energy Engineering Group Co., Ltd., Beijing, China\\
$^{5}$Kiel University, Kiel, Germany
}

\begingroup
\renewcommand{\thefootnote}{}
\footnotetext{$^\dagger$ Corresponding author: \texttt{liumengyuan@pku.edu.cn}.}
\endgroup

\maketitle

\begin{abstract}
Millimeter-wave (mmWave) radar provides privacy-preserving sensing and is valuable for human action recognition (HAR). Existing mmWave point cloud datasets are limited in scale and mostly collected under homogeneous single-source settings, preventing current methods from handling real-world distribution shifts caused by heterogeneous radar sources, such as different devices and frequency bands. To address this, we introduce UniMM-HAR, the largest and first mmWave point cloud HAR dataset for heterogeneous multi-source scenarios, standardizing three distinct radar configurations to realistically evaluate cross-source generalization. We further propose the Doppler-aware Point Cloud Network (DAP-Net) to tackle heterogeneity challenges. DAP-Net enhances intra-modal representations and performs cross-modal alignment to learn source-invariant action semantics. Leveraging action-consistent spatio-temporal Doppler patterns as anchors, the Dual-space Doppler Reparameterization (D²R) module performs sample-adaptive geometric densification and Doppler-guided feature recalibration, while the Text Alignment Module (TAM) provides stable semantic anchors via a pretrained textual space. Experiments show that DAP-Net significantly outperforms existing methods under heterogeneous radar settings, achieving state-of-the-art accuracy and strong cross-source robustness.
Code and dataset are publicly available at \href{https://github.com/jolin830/DAP-Net}{\textcolor{blue}{https://github.com/jolin830/DAP-Net}}.
    \keywords{Radar Point Cloud \and Heterogeneous mmWave Action Recognition  \and mmWave Doppler}
\end{abstract}

\section{Introduction}
\label{sec:intro}

Human action recognition (HAR) is fundamental to video understanding and intelligent homes \cite{feichtenhofer2019slowfast,liu2017enhanced,11278750,hinojosa2022privhar,lin2024sfmvit}. Most existing approaches rely on camera-based modalities, including RGB \cite{duan2022revisiting,bruce2022mmnet}, 
and skeleton \cite{yan2018spatial, Chen_2021_ICCV, 10113233,liu20253d,liu2024multi,yin2025recognizing,liu2024hdbn}. These methods inevitably expose identifiable visual information and remain highly sensitive to illumination variations and occlusion, which constrains their deployment in privacy-sensitive environments \cite{xia2024timestamp,xu2025ai}. In contrast, millimeter-wave (mmWave) radar captures human motion through electromagnetic reflections rather than appearance cues, thereby inherently preserving privacy while offering superior robustness to low-light conditions and visual obstructions, making it more suitable for continuous and unobtrusive monitoring \cite{palipana2021pantomime,salami2022tesla,liu2025revealing,wan2014gesture}.

\begin{figure}[!t]
    \centering
    \includegraphics[width=1\linewidth]{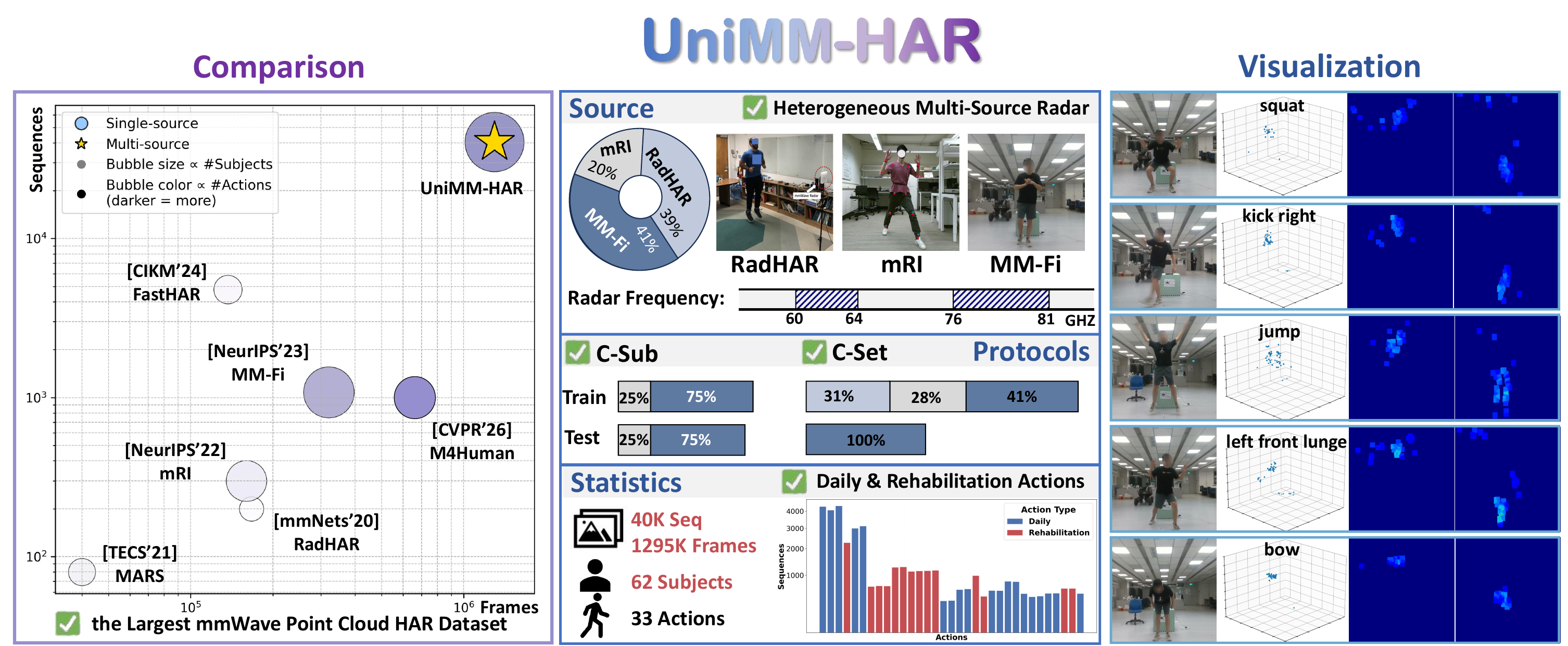}
    \caption{
    UniMM-HAR is currently the largest mmWave point cloud human action recognition dataset and the first unified benchmark for heterogeneous multi-source distributions, providing Cross-Subject and Cross-Set evaluation protocols and covering both daily and rehabilitation actions. The samples in UniMM-HAR are collected from radar devices with different models and operating frequencies.
     }
    \label{fig:Intro}
\end{figure}

In recent years, mmWave-based Human action recognition has advanced rapidly, largely driven by public radar datasets, with point cloud datasets being particularly valuable due to their explicit 3D structure and rich geometric semantics \cite{mmfi,m_activity,mRI,shao2024fast,singh2019radhar,fastHAR}.
However, existing mmWave point cloud HAR datasets are mostly limited in scale and collected under homogeneous single-source settings, where source refers to radar configurations such as device model and operating frequency band. Accordingly, current methods are primarily designed and validated under this simplified assumption. In contrast, real-world mmWave sensing scenarios are typically heterogeneous and multi-source, as different applications may deploy diverse radar devices, and practical settings often involve device upgrades and multi-device collaboration. Therefore, a practically deployable HAR system must be capable of adapting to distribution shifts arising from different radar sources.
The core challenge of heterogeneous multi-source data lies in the fact that the same human action, when captured by different radar sources, can exhibit significant distribution shifts in terms of point density, spatial distribution, velocity statistics, and noise characteristics. These shifts are not merely additive noise but structural variations tightly coupled with device-specific properties. Models trained on single-source data tend to overfit source-specific statistical patterns rather than learning stable, source-agnostic action semantics, leading to substantial performance degradation under distribution changes and hindering deployment in multi-source environments \cite{torralba2011unbiased,liu2024unbiased,zhou2022domain}. Therefore, it is necessary to \uline{construct a large-scale mmWave point cloud dataset that covers heterogeneous multi-source distributions and to explore effective modeling strategies tailored to such heterogeneous settings}, in order to enhance cross-source generalization capability.

Motivated by this, \uline{we introduce UniMM-HAR, a large-scale heterogeneous radar point cloud HAR dataset, along with the Doppler-aware Point Cloud Network (DAP-Net)}.
To our knowledge, UniMM-HAR is the largest mmWave point cloud HAR dataset and the first unified benchmark from heterogeneous multi-source data as Fig.\ref{fig:Intro}. It standardizes data from three heterogeneous radar configurations and provides cross-subject (C-Sub) and cross-set (C-Set) protocols to evaluate generalization under inherent inter-source distribution shifts. Compared with single-source benchmarks, UniMM-HAR more rigorously tests whether models learn source-agnostic and semantically meaningful representations.
To address heterogeneous distribution shifts, DAP-Net performs intra-modal enhancement and cross-modal alignment. It leverages the action-consistent spatio-temporal Doppler patterns and introduces the Dual-space Doppler Reparameterization (D²R) module to decouple source-specific interference in geometric and feature spaces. Furthermore, a lightweight Text Alignment Module (TAM) aligns mmWave features with a pretrained textual semantic space, serving as a stable semantic anchor to improve cross-source robustness.

DAP-Net is built upon the D²R module, which consists of Doppler-guided Geometry Reparameterization (DGR) and Motion-aware Feature Recalibration (MFR) for geometric and feature-level modeling, respectively. The reparameterized point clouds are processed by a point cloud backbone, and the extracted features are further aligned via TAM.
In the geometric space, given the sparsity and noise of mmWave point clouds, DGR performs Doppler-guided, sample-adaptive densification over motion-relevant regions while suppressing device-induced noise and outliers, instead of applying a fixed processing strategy. In the feature space, MFR dynamically recalibrates feature channels conditioned on Doppler cues, enhancing motion-related responses and mitigating source-specific bias. This sample-wise reparameterization enables more consistent cross-source representations and is implemented as a parameter-efficient, plug-and-play module compatible with various backbones.
To further enhance semantic consistency, TAM projects mmWave features into a pretrained textual semantic space constructed from action prompts and aligns them via similarity-based learning, serving as a stable semantic anchor for cross-source robustness.

In summary, this work has the following contributions: 
\begin{itemize}
    \item We construct and release UniMM-HAR, the largest mmWave point cloud HAR dataset to date and the first unified benchmark designed for heterogeneous multi-source distributions. It more realistically simulates multi-source deployment scenarios in real-world applications and provides a foundation for cross-source generalization research.
    \item We propose DAP-Net for cross-source adaptation of heterogeneous mmWave point clouds via intra-modal and cross-modal enhancement. The Dual-space Doppler Reparameterization (D²R) module performs sample-adaptive geometric densification and feature recalibration to mitigate source-specific biases, while the Text Alignment Module (TAM) provides stable semantic anchors to improve semantic  consistency.
    \item DAP-Net achieves state-of-the-art performance on UniMM-HAR, demonstrating its effectiveness and cross-source robustness. Notably, the D²R module is a parameter-efficient, plug-and-play design that can seamlessly integrate with any point cloud backbone, enhancing mmWave modeling.
\end{itemize}

\section{Related Work}
\label{sec:relatedwork}
\subsection{Human Action Recognition on Point Clouds}
Point clouds from LiDAR or depth cameras have led to mature point cloud networks, typically categorized into static and dynamic paradigms. Static point clouds excel in tasks such as object classification and HAR \cite{qi2017pointnet,ma2022rethinking,luo2023edgeactnet,kim2021point,qi2017pointnet++,xu2024pointllm,deng2024vg4d}, while dynamic point clouds model temporal motion \cite{ben20243dinaction,pst_transformer,ustssm,liu2025mamba4d}. Although dense point clouds provide fine-grained motion cues, they incur high hardware and computational costs . In contrast, mmWave radar offers a low-cost, privacy-preserving alternative. However, its sparse and noisy nature prevents direct adoption of dense point cloud architectures. DAP-Net addresses this limitation by incorporating Doppler-guided motion priors to compensate for geometric sparsity, enabling dense point cloud backbones to effectively process radar point clouds.

\begin{table*}[!t]
\centering
\caption{Comparison of UniMM-HAR with existing mmWave point cloud HAR datasets. 
For fair comparison, only the single-person subset of mmMulti \cite{zhou2025mmmulti} is evaluated. 
RF-BW: radar frequency band and bandwidth (GHz). 
Multi-Source: whether the dataset is collected from heterogeneous radar source. 
$^\dagger$: non-public data. 
*: originally designed for other tasks.
Source datasets of UniMM-HAR are in light gray.}
\label{tab:dataset_comparison}
\setlength{\tabcolsep}{4pt}
\renewcommand{\arraystretch}{1.2}
\resizebox{\linewidth}{!}{
\begin{tabular}{l | cccc | ccc | cc | cc}
\toprule

\multirow{2}{*}{\textbf{Dataset}} 
& \multicolumn{4}{c}{\textbf{Statistics}} 
& \multicolumn{3}{c}{\textbf{Radar}} 
& \multicolumn{2}{c}{\textbf{Actions}} 
& \multicolumn{2}{c}{\textbf{Protocols}} \\

\cmidrule(lr){2-5}
\cmidrule(lr){6-8}
\cmidrule(lr){9-10}
\cmidrule(lr){11-12}

& \# Act. & \# Sub. & \# Seq. & \# Frame
& Device & RF-BW & Multi-Source    
& Daily & Rehab
& C-Sub & C-Set \\

\midrule

\cellcolor{sourcegray} RadHAR~\cite{singh2019radhar} 
& 5 & 2 & 200 & 167k
& TI IWR1443 & 76--81 & \xmark
& \cmark & \xmark
& \xmark & \xmark \\

$^*$MARS~\cite{an2021mars} 
& 10 & 4 & 80 & 40k
& TI IWR1443 & 76--81 & \xmark
& \cmark & \xmark
& \xmark & \xmark \\

m-Activity$^\dagger$~\cite{m_activity} 
& 5 & 9 & N/A & 15k
& TI IWR1443 & 76--81 & \xmark
& \cmark & \xmark
& \xmark & \xmark \\

\cellcolor{sourcegray} mRI~\cite{mRI} 
& 12 & 20 & 300 & 160k
& TI IWR1443 & 76--81 & \xmark
& \cmark & \cmark
& \cmark & \xmark \\

miliPoint~\cite{cui2023milipoint} 
& 49 & 11 & N/A & 545k
& TI IWR1843 & 76--81 & \xmark
& \cmark & \xmark 
& \xmark & \xmark \\

\cellcolor{sourcegray} MM-Fi~\cite{mmfi} 
& 27 & 40 & 1080 & 320k
& TI IWR6843 & 60--64 & \xmark
& \cmark & \cmark
& \cmark & \cmark \\

$^*$mmParse$^\dagger$~\cite{mmparse} 
& 10 & 32 & N/A & N/A
& TI IWR6843 & 60--64 & \xmark
& \cmark & \xmark
& \xmark & \xmark \\

FastHAR~\cite{fastHAR} 
& 7 & 5 & 4776 & 137k
& TI IWR6843 & 60--64 & \xmark
& \cmark & \xmark
& \xmark & \xmark \\

$^*$milliflow~\cite{ding2024milliflow} 
& 7 & 12 & N/A & 43k
& Vayyar vTrigB & 62--69 & \xmark
& \cmark & \xmark
& \cmark & \xmark \\

mmMulti$^\dagger$~\cite{zhou2025mmmulti} 
& 6 & 4 & 2520 & N/A
& TI IWR1443 & 76--81 & \xmark
& \cmark & \xmark
& \cmark & \cmark \\

$^*$M4Human~\cite{fan2026m4human}  
& 50 & 20 & 999 & 661k
& Vayyar vTrigB & 62--69 & \xmark
& \cmark & \cmark
& \cmark & \xmark \\

\midrule

\rowcolor{oursgray}
\textbf{UniMM-HAR}
& 33 & \textbf{62} & \textbf{40494} & \textbf{1295k}
& \makecell{TI IWR1443 \\ \& TI IWR6843}
& \makecell{76--81 \\ \& 60--64}
& \textbf{\cmark}
& \cmark & \cmark
& \cmark & \cmark \\

\bottomrule
\end{tabular}
}
\end{table*}

\subsection{mmWave-based Human Action Recognition}
Millimeter-wave action recognition has witnessed substantial progress with diverse data representations, predominantly spectrum-based \cite{rd_1,rd_2,rd_3,ra_2,microd_1,microd_2,microd_3,wang2016interacting,fhager2019pulsed} and point cloud–based approaches \cite{lai2026radarllm,vox_1,vox_2,vox_3,singh2019radhar,mmw_1,mmw_2,mmw_3,qi2017pointnet}. 
Spectrum-based methods represent radar signals as 2D spectrum maps and apply image recognition models to capture Doppler cues, but the 2D projection inevitably causes information loss.
Point cloud-based methods preserve 3D structure but often overlook Doppler information, a distinctive characteristic of mmWave sensing. Existing studies mainly use Doppler for preprocessing, such as frame selection \cite{guo2023point}, rather than explicit motion representation learning. In contrast, DAP-Net leverages Doppler to guide feature learning in both geometric and feature spaces.

\subsection{Existing mmWave Radar Datasets}

The advancement of mmWave-based human understanding has been largely facilitated by public radar datasets \cite{chen2022mmbody,mmgait,xue2021mmmesh,an2021mars,sengupta2022mmpose,lee2023hupr,choi2025mvdopplerpose,fan2026m4human}, with action recognition gaining increasing attention \cite{zhao2023cubelearn, hor2023mvdoppler, zhang2018cnn, mmfi, singh2019radhar, m_activity, mRI, cui2023milipoint, ding2024milliflow, fastHAR, zhou2025mmmulti, zhao2018rf}. Point cloud datasets are increasingly favored for their intuitive 3D structure and rich geometric semantics. Early mmWave point cloud HAR datasets \cite{zhao2018rf, singh2019radhar, an2021mars, m_activity} were limited in scale and relied on random splits. Later datasets \cite{mRI, cui2023milipoint, mmfi, fastHAR, zhou2025mmmulti} expanded subjects and actions and introduced cross-subject (C-Sub) and cross-dataset (C-Set) evaluation protocols. Existing datasets remain single-source and limited in scale, diversity, and standardization. UniMM-HAR addresses these gaps by integrating heterogeneous multi-source radar data, significantly increasing samples, subjects, and action diversity, and providing multiple standardized evaluation protocols under heterogeneous distributions.

\section{UniMM-HAR Dataset}

\subsection{Dataset Overview}

Existing mmWave point cloud datasets suffer from limitations in scale, subject diversity, and action coverage. They are typically collected using a single radar device at a fixed frequency band and lack standardized evaluation protocols. Inspired by \cite{mahmood2019amass}, we adopt a large-scale unified paradigm that integrates heterogeneous mmWave point cloud sources under standardized evaluation protocols, resulting in the proposed UniMM-HAR. Table \ref{tab:dataset_comparison} provides a systematic comparison with existing datasets.
Specifically, UniMM-HAR consolidates three representative datasets: RadHAR \cite{singh2019radhar}, mRI \cite{mRI}, and MM-Fi \cite{mmfi}, and standardizes their action categories, subject splits, and sample representations to form a unified benchmark. \uline{The heterogeneity in UniMM-HAR mainly arises from differences in radar devices, operating frequency bands, and other recording configs across the sub-datasets.} These differences lead to varying data distributions even for the same action, providing a realistic and challenging setting for cross-domain evaluation. UniMM-HAR contains 33 action categories spanning daily and rehabilitation activities, 62 subjects, and over 1.29 million frames. Figure \ref{fig:dataset_visulization} visualizes representative action samples. To the best of our knowledge, UniMM-HAR is the largest mmWave point cloud dataset for HAR to date and the first to systematically support cross-subject and cross-set evaluation under heterogeneous radar hardware and frequency bands.

\subsection{Dataset Statistics and Unification Process}\label{sec:SourceDatasets}
\begin{figure}[t]
    \centering
    \includegraphics[width=1\linewidth]{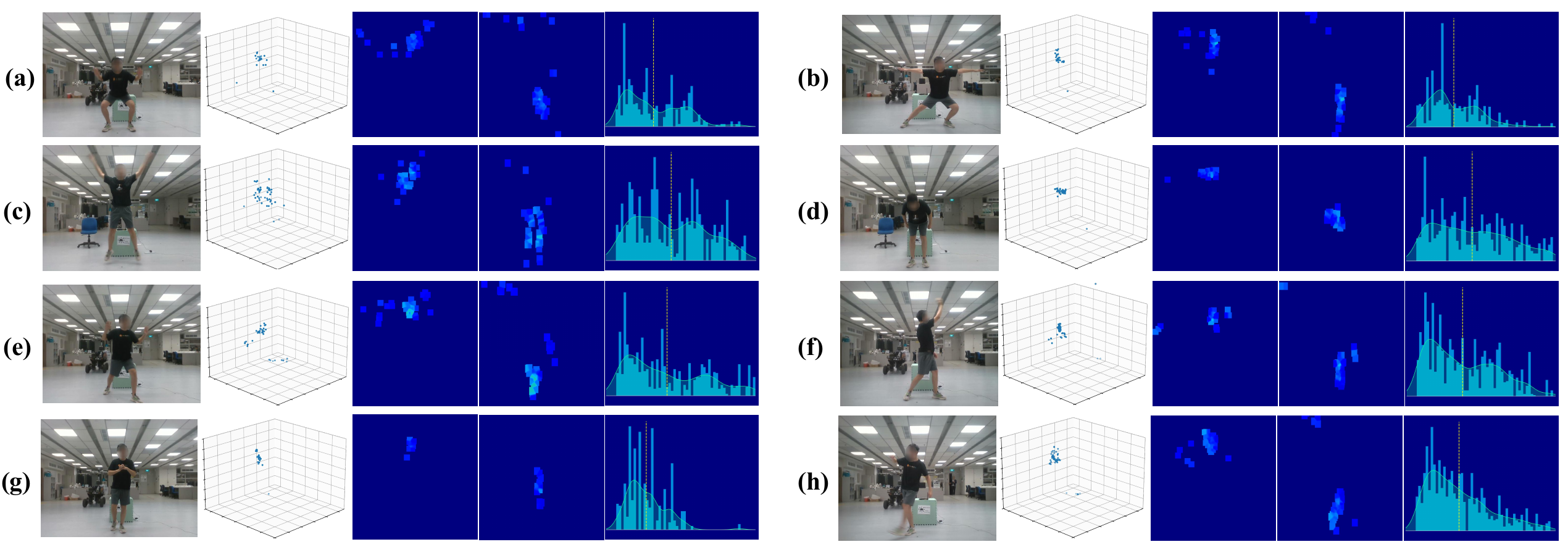}
    \caption[UniMM-HAR actions]{Visualization of different action samples in UniMM-HAR. Actions (a)–(h) are: \textit{squat, left lunge,  jump, bow, left front lunge, throw left, stretch, and kick right}. In each subfigure, the 1st and 2nd columns show the RGB video frame and mmWave point cloud, respectively. The 3rd and 4th columns show the Doppler heatmap in BEV and front view. The 5th column shows the Doppler distribution for the action. 
    }
    \label{fig:dataset_visulization}
\end{figure}

\begin{figure*}[t]
    \centering
    \begin{minipage}[t]{0.66\linewidth}
        \vspace{-4mm} 
        \centering
        \small
        \setlength{\tabcolsep}{4pt}
        \captionof{table}{
        Dataset statistics before and after processing. 
 The processing strategy (Proc.): Sliding Window 
 (SW), Segmentation (SEG), and Normalization (Norm).
        }
        \resizebox{\linewidth}{!}{
        \begin{tabular}{l l l r r r r}
        \toprule
        \textbf{Dataset} & \textbf{Proc.} & \textbf{Format} & \textbf{\#Act.} & \textbf{\#Sub.} & \textbf{\#Seq.} & \textbf{\#Frame(k)} \\
        \midrule
        \multirow{3}{*}{RadHAR \cite{singh2019radhar}} & -- & TXT & 5  & 2 & 200    & 167 \\
                                & SW  & CSV & 5  & 2 & 15,715 & 942 \\
                                & SW  & NPZ & 5  & 2 & 15,715 & 502 \\
        \multirow{3}{*}{mRI \cite{mRI}}    & -- & CSV & 12 & 20 & 300    & 160 \\
                                & SW  & CSV & 12 & 20 & 8,332  & 131 \\
                                & SW  & NPZ & 12  & 20 & 8,332 & 266 \\
        \multirow{3}{*}{MM-Fi \cite{mmfi}}  & -- & BIN & 27 & 40 & 1,080  & 320 \\
                                & SEG & CSV & 27 & 40 & 16,447 & 310 \\
                                & SEG & NPZ & 27 & 40 & 16,447 & 526 \\
        \midrule
        \multirow{2}{*}{\textbf{UniMM-HAR}} & \textbf{--}  & \textbf{CSV} & \textbf{33} & \textbf{62} & \textbf{40,494} & \textbf{1,384} \\
                                         & \textbf{Norm} & \textbf{NPZ} & \textbf{33} & \textbf{62} & \textbf{40,494} & \textbf{1,295} \\
        \bottomrule
        \end{tabular}}
        \label{tab:dataset_processing}
    \end{minipage}%
    \hfill
    \begin{minipage}[t]{0.32\linewidth}
        \vspace{0pt} 
        \centering
        \includegraphics[width=0.95\linewidth]{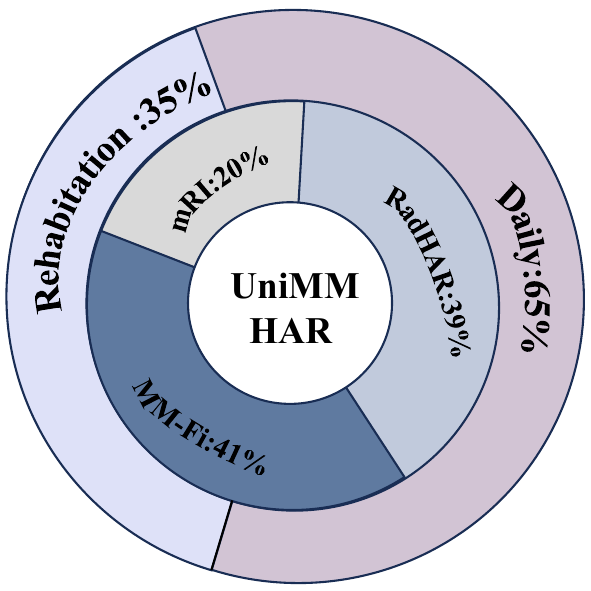}
        \captionof{figure}{Action distribution across sources and types. }
        \label{fig:dataset_distribution}
    \end{minipage}
\end{figure*}

UniMM-HAR excludes datasets that are not publicly available, incompletely annotated, or lacking raw point cloud data, and ultimately retains three high-quality datasets that satisfy the unification criteria: RadHAR \cite{singh2019radhar}, mRI \cite{mRI}, and MM-Fi \cite{mmfi}. The selected datasets are then systematically unified. Table \ref{tab:dataset_processing} summarizes the statistics of the source datasets and UniMM-HAR before and after processing. The procedure is detailed as follows.

\noindent \textbf{Action Alignment}. After consolidating synonymous actions across the three datasets, UniMM-HAR establishes a unified taxonomy of 33 categories, including 21 daily activities and 12 rehabilitation actions as in Fig.\ref{fig:dataset_distribution}. The action distribution follows a natural long-tailed pattern, preserving realistic class imbalance while increasing the difficulty of the benchmark.

\noindent \textbf{Dataset-Aware Preprocessing}. Given substantial differences among datasets in sequence length, annotation granularity, and structural organization, dataset-aware preprocessing strategies are applied to generate standardized clips. RadHAR is processed using sliding windows (window size 60, stride 10), consistent with its official implementation. mRI adopts sliding windows (window size 32, stride 16).
Since MM-Fi sequences are not pre-segmented by action, they are divided into valid action clips according to the provided segmentation files.

\noindent \textbf{Format Standardization}. To improve interoperability and reproducibility, UniMM-HAR is released in two formats: CSV and NPZ. The CSV format preserves raw annotations for data provenance and traceability, while NPZ is optimized for direct model input. All samples follow a unified naming convention encoding dataset source, action label, subject identity, and scene.

\noindent \textbf{Representation Standardization}. Temporal–point normalization converts each clip to a fixed shape $[T, P, C] = [32, 64, 5]$, with channels $x, y, z$, Doppler, and intensity.
When the original sequence length exceeds $T$, uniform temporal downsampling is applied. Otherwise, zero-padding is used. When the number of points per frame exceeds $P$, Farthest Point Sampling (FPS) is applied. Otherwise, repeat sampling is performed. 
Furthermore, UniMM-HAR offers dataset-level and clip-level normalization variants. These normalize $x, y, z$, Doppler, and intensity using global or per-sample statistics to reduce severe statistical bias, while retaining multi-source distribution gaps as an essential challenge of the benchmark.

\subsection{Evaluation Protocols}
\label{sec:Evaluation}
UniMM-HAR supports three multi-source evaluation protocols: Random Split, Cross-Subject (C-Sub), and Cross-Set (C-Set).
1) Random Split partitions samples 60:40 to simulate seen-subject and seen-action scenarios. 
2) Cross-Subject uses mRI and MM-Fi, comprising 19,657 samples across 25 actions and 60 subjects, with a 5:5 split by subjects. 
3) Cross-Set covers 28,041 sequences across 27 actions in 6 scenes, following a 4:2 split by scene for training and testing.
Importantly, all evaluation protocols are conducted under the heterogeneous radar configuration, which introduces additional domain shifts across sources. As a result, these settings are more challenging than the conventional C-Sub and C-Set protocols defined under single-source radar setups.

\begin{figure}
    \centering
    \includegraphics[width=1\linewidth]{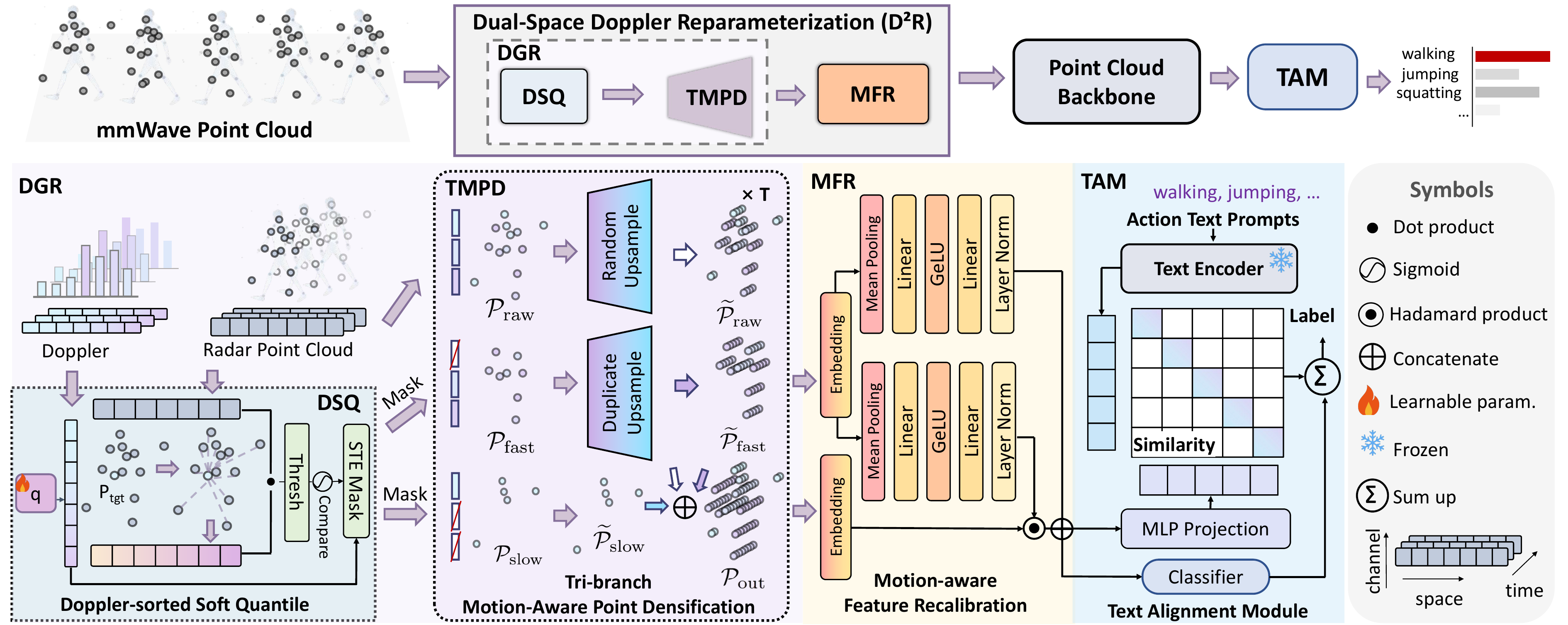}
    \caption{\textbf{Overview of DAP-Net:} 
    First, the Dual-space Doppler Reparameterization (D²R) converts mmWave point cloud sequences into Doppler-guided dense representations. Within D²R, Doppler-guided Geometry Reparameterization (DGR) performs geometric densification, while Motion-aware Feature Recalibration (MFR) enhances motion-sensitive feature modeling to produce point embeddings. 
    These embeddings are first processed by a point cloud backbone to extract global features and then aligned with textual features through the Text Alignment Module (TAM), with final predictions made based on feature similarity.
    }
    \label{fig:DAP}
\end{figure}

\section{Methodology}

\subsection{Task Definition}
We consider a mmWave point cloud sequence represented as 
$X \in \mathbb{R}^{T \times P \times 3} = \{\mathcal{P}_t\}_{t=1}^{T}$,
where $T$ denotes the frame number, $P$ denotes the point number per frame, 
and each point is represented by its 3D coordinates $(x, y, z)$. 
The Doppler velocity of each point is represented by 
$v \in \mathbb{R}^{T \times P}$. 
The point number in frame $t$ is denoted by $P_t = |\mathcal{P}_t|$. 
The objective is to predict the action label $y$ associated with the input sequence.
An overview of our DAP-Net is shown in Fig.~\ref{fig:DAP}.

\subsection{Dual-Space Doppler Reparameterization (D$^2$R)}
To effectively address distribution shifts from heterogeneous multi-source data, it is essential to disentangle source-specific representations from action-related ones and extract motion priors that remain stable across different radar sources. Geometric density and absolute coordinates are highly dependent on hardware configurations, whereas Doppler encodes the radial velocity of each point, directly reflecting the underlying physical motion \cite{chae2025doppler,haitman2025doppdrive,prabhakara2023high,liu2023echoes}. Although absolute Doppler values may vary across devices, the relative temporal evolution exhibits consistent patterns across sources. For example, arm swinging produces periodic velocity oscillations, whereas squatting shows a regular sign alternation of velocity across the action cycle. Therefore, action-consistent spatio-temporal Doppler patterns can serve as reliable modeling anchors.

Based on this observation, D²R leverages Doppler information as a motion prior and performs sample-adaptive modeling via two complementary modules: Doppler-guided Geometric Reparameterization (DGR), which densifies motion-critical points in the geometric space, and Motion-aware Feature Recalibration (MFR), which conditionally recalibrates dense point cloud features in the feature space to enhance sensitivity to dynamic patterns. Notably, the D²R module is parameter-efficient and plug-and-play, enabling flexible integration with various point cloud backbone networks while enhancing mmWave point cloud modeling with negligible additional computational cost.

\subsection{Doppler-guided Geometric Reparameterization (DGR)}\label{sec:DGR}

Millimeter-wave point clouds are inherently sparse and noisy, with distribution discrepancies further amplified under heterogeneous multi-source settings, rendering fixed-rule densification strategies designed for single-source scenarios difficult to generalize: 1) Repetitive upsampling indiscriminately duplicates both informative motion points and noise, amplifying cross-source bias. 2)  Sliding-window aggregation relies on manually defined temporal spans and remains sensitive to variations in frame rate and point density.
To better address densification and denoising under heterogeneous distributions, DAP-Net introduces a Doppler-guided Geometry Reparameterization (DGR) module, which performs selective densification of each frame to a fixed target size $P_{\text{goal}}$ using per-point Doppler as a motion prior. By adaptively emphasizing discriminative motion regions and suppressing noise, DGR enables heterogeneous point clouds to form aligned spatial structures, thereby improving cross-source generalization.

\noindent \textbf{Doppler-sorted Soft Quantile (DSQ)}
Unlike methods relying on fixed absolute thresholds, DSQ introduces a learnable quantile parameter $q \in (0,1)$ to characterize the relative motion intensity boundary of each frame, enabling adaptive partitioning based on the relative distribution.
For each frame $\mathcal{P}_t$, let the Doppler magnitudes be $\{v_t^i\}_{i=1}^{P_t}$, sorted in ascending order as
$v_t^1 \le \cdots \le v_t^{P_t}$.

The target rank position for frame $t$ is defined as
$P_t^{\text{tgt}} = q(P_t - 1)$,
representing the desired quantile index. To obtain a differentiable approximation of this rank, Gaussian weights are constructed centered at $P_t^{\text{tgt}}$:
\begin{equation} \label{eq:wti}
w_t^i \propto
\exp\!\left(
-\frac{\left(i - P_t^{\text{tgt}}\right)^2}{2\sigma^2}
\right).
\end{equation}

The frame-level motion threshold is computed as
$\tau_t = \sum_{i=1}^{P_t} w_t^i\, v_t^i$,
which can be interpreted as a soft dot product between the sorted Doppler values and the Gaussian weight distribution.
Given the threshold $\tau_t$, each Doppler value $v_t^i$ is compared against it to obtain a soft motion indicator:
\begin{equation} \label{eq:sti}
s_t^i = \sigma\!\left(
\frac{v_t^i - \tau_t}{\gamma}
\right),
\end{equation}
where $\gamma$ controls the sharpness of the transition. The resulting scores are then binarized using a Straight-Through Estimator (STE)~\cite{bengio2013estimating,courbariaux2016binarized}, maintaining explicit partitioning in the forward pass while propagating gradients through the soft probabilities $s_t^i$ in backpropagation.
By modeling relative statistics via quantiles, DSQ achieves stable motion-aware partitioning under heterogeneous multi-source distributions, effectively mitigating source-induced statistical shifts and providing consistent motion priors for subsequent geometric reparameterization.

\noindent \textbf{Tri-branch Motion-Aware Point Densification (TMPD)}
Given the motion confidence $s_i$ for each point, TMPD partitions the point set into three complementary subsets:
$
\mathcal{P}_{\text{fast}} = \{\mathbf{p}_i \mid s_i > \delta\}, \quad
\mathcal{P}_{\text{slow}} = \{\mathbf{p}_i \mid s_i \le \delta\}, \quad
\mathcal{P}_{\text{raw}} = \mathcal{P},
$
where $\delta$ is the partition threshold and $\mathcal{P}$ denotes the original point set. The points are assigned to Fast, Slow, and Raw branches, which are processed as:
\begin{equation} \label{eq:P_process}
\widetilde{\mathcal{P}}_{\text{fast}}
= \bigcup_{k=1}^{r} \mathcal{P}_{\text{fast}}, \quad
\widetilde{\mathcal{P}}_{\text{slow}}
= \mathcal{P}_{\text{slow}}, \quad
\widetilde{\mathcal{P}}_{\text{raw}}
= \operatorname{RandSample}\big(\mathcal{P}_{\text{raw}}, P_{\text{raw}}\big),
\end{equation}
where $r$ is the duplication factor for motion-salient points in the Fast branch, and $\operatorname{RandSample}(\cdot, N)$ uniformly samples $N$ points for the Raw branch.  
The \textbf{Fast branch} focuses on motion-salient regions. Since mmWave point clouds are inherently sparse in high-motion areas and exhibit distribution discrepancies across heterogeneous sources, $r$-fold controlled duplication increases the sampling density of dynamic structures across sources, improving cross-source representation consistency. Generative synthesis or interpolation is avoided to prevent physically inconsistent noise due to the lack of point-wise supervision and stable inter-frame correspondences.  
The \textbf{Slow branch} handles quasi-static points, which may belong to low-motion human regions or background structures. Their geometry is relatively stable, so the original distribution is preserved without densification to avoid cross-source geometric shifts affecting feature consistency.  
For the \textbf{Raw branch}, the number of points is 
$\widetilde{P}_{\text{raw}} = P_{\text{goal}} - r P_{\text{fast}} - P_{\text{slow}}$, ensuring the final point cloud has cardinality $P_{\text{goal}}$.
The final densified point cloud $\mathcal{P}_{\text{out}}$, with cardinality $|\mathcal{P}_{\text{out}}| = P_{\text{goal}}$, is obtained by aggregating the three branches:
\begin{equation} \label{eq:Pgoal}
\mathcal{P}_{\text{out}}
= \widetilde{\mathcal{P}}_{\text{fast}}
\cup
\widetilde{\mathcal{P}}_{\text{slow}}
\cup
\widetilde{\mathcal{P}}_{\text{raw}},
\end{equation}
This branch-wise adaptive strategy enhances key motion information while mitigating sparsity and noise discrepancies from heterogeneous sources.

\subsection{Motion-aware Feature Recalibration (MFR)}
\label{sec:MFR}

Previous works primarily focus on modeling the geometric 3D structure of point clouds, paying limited attention to the motion information embedded in Doppler signals. This results in extracted features with constrained ability to capture dynamic motion patterns. To explicitly enhance motion representation, DAP-Net introduces the Motion-aware Feature Recalibration (MFR) module, which conditionally recalibrates dense point cloud features using Doppler-guided fast points, thereby maintaining consistent motion information across heterogeneous radar sources.
The dense point clouds $\mathcal{P}_{\text{out}} \in \mathbb{R}^{T \times P_{\text{out}} \times 3}$ and the fast points $\mathcal{P}_{\text{fast}} \in \mathbb{R}^{T \times P_{\text{fast}} \times 3}$ are embedded to obtain $\mathcal{F}_{\text{out}} \in \mathbb{R}^{T \cdot P_{\text{out}} \times C}$ and $\mathcal{F}_{\text{fast}} \in \mathbb{R}^{T \cdot P_{\text{fast}} \times C}$, respectively.
The fast point features $\mathcal{F}_{\text{fast}}$  are then aggregated into a motion summary vector $c \in \mathbb{R}^{C}$ to guide subsequent feature recalibration:
\begin{equation}
    c = \frac{1}{|\mathcal{F}_{\text{fast}}|} \sum_{i \in \mathcal{F}_{\text{fast}}} f_i,
\end{equation}
where $f_i$ denotes the embedding of the $i$-th fast point. This vector provides a stable motion prior across heterogeneous radar sources.
Inspired by~\cite{perez2018film}, the MFR module leverages $c$ to perform channel-wise recalibration of $\mathcal{F}_{\text{out}}$. 
Specifically, MLPs generate learnable scaling $\gamma$ and shifting $\beta$ parameters from $c$:
\begin{equation}
    \gamma = \text{MLP}_{\gamma}(c), \quad
\beta = \text{MLP}_{\beta}(c), \quad
\gamma, \beta \in \mathbb{R}^{C}.
\end{equation}
Recalibrated dense features $\mathcal{F} $ are derived via channel-wise affine transformation:
\begin{equation}
    \mathcal{F} = \gamma \odot \mathcal{F}_{\text{out}} + \beta,
\end{equation}
where $\odot$ denotes element-wise multiplication across channels. 
This operation leverages the motion-guided fast points to modulate point cloud features, 
effectively enhancing the network's sensitivity to motion patterns 
while maintaining consistent motion representation across heterogeneous multi-source scenarios.

\subsection{Backbone and Text Alignment Module}
\label{sec:backbone_and_cam_en}

DAP-Net is backbone-agnostic and can be flexibly integrated into various point cloud backbone networks.
Given the input point cloud features
$
\mathcal{F} \in \mathbb{R}^{T \cdot P_{\text{out}} \times C},
$
the backbone network extracts high-level representations, which are aggregated via max pooling to obtain a global feature:
$
\tilde{\mathcal{F}} = \text{MaxPool}(\text{Backbone}(\mathcal{F})) \in \mathbb{R}^{D}.
$
This global feature \(\tilde{\mathcal{F}}\) is fed into a classifier to obtain action prediction logits
$
z \in \mathbb{R}^{K},
$
where \(K\) denotes the number of action classes.

mmWave point clouds are sparse and have lower point resolution. 
To leverage this, DAP-Net introduces the \textbf{Text Alignment Module (TAM)}, aligning mmWave features with information-rich textual embeddings.  
TAM first encodes prompts $T(\cdot)$ for all action categories using a pretrained text encoder \(E_{\text{text}}(\cdot)\):
\begin{equation}
    F_{\text{text}} = E_{\text{text}}\big([\text{T(text}_i)]_{i=1}^{K}\big) \in \mathbb{R}^{K \times C_{\text{text}}},
\end{equation}
where \(C_{\text{text}}\) denotes the dimensionality of the text embeddings.
The global mmWave feature $\tilde{\mathcal{F}}$ is projected via an MLP into the shared textual embedding space, yielding the feature $F_{\text{mmw}} \in \mathbb{R}^{K \times C_{\text{text}}}$. Then, TAM computes the similarity between mmWave and text features:
\begin{equation}
    S = \text{Softmax}\big(F_{\text{mmw}} F_{\text{text}}^\top \big) \in \mathbb{R}^{K \times K}.
\end{equation}
Finally, the similarity scores $s=diag(S)\in \mathbb{R}^{K}$ are fused with the backbone classifier logits $z \in \mathbb{R}^{K}$ to produce the final prediction $\hat{\mathbf{y}}\in \mathbb{R}^{K}$.
Note that TAM is independent of the specific text encoder and can be seamlessly replaced with more advanced large language models (LLMs) or domain-adapted text encoders.

\begin{table*}[t]
\footnotesize
\centering
\setlength{\tabcolsep}{10pt}
\caption{Comparison with prior methods on UniMM-HAR C-Sub and C-Set. 
Original PC type indicates the point cloud type the model was designed for. 
MPC: mmWave radar point clouds; 
SPC: static point clouds; 
TPC: temporal point clouds.}
\label{tab:sota_comparison}

\begin{tabular}{l r c r r}
\toprule
\textbf{Model} & \textbf{Source} & \textbf{Original } 
& \multicolumn{2}{c}{\textbf{UniMM-HAR (\%)}} \\
\cmidrule(lr){4-5}
& &\textbf{PC Type} & \textbf{C-Sub} & \textbf{C-Set} \\
\midrule
PointNet \cite{qi2017pointnet} & CVPR'17 & SPC & 59.27 & 70.96 \\
DGCNN \cite{phan2018dgcnn} & NN'18 & SPC & 71.70 & 52.18 \\
RadHAR \cite{singh2019radhar} & mmNSS'19 & MPC & 42.49 & 48.47 \\
PointMLP \cite{ma2022rethinking} & ICLR'22 & SPC & 71.06 & 78.13 \\
PST-Transformer \cite{pst_transformer} & TPAMI'22 & TPC & 63.13 & 79.70 \\
PointCLIP \cite{zhang2022pointclip} & CVPR'22 & SPC & 67.82 & 69.82 \\ 
Clip2point \cite{huang2023clip2point} & ICCV'23 & SPC & 64.17 & 59.56 \\    
FastHAR \cite{shao2024fast} & CIKM'24 & MPC & 53.72 & 61.16 \\
3DInAction \cite{ben20243dinaction} & CVPR'24 & TPC & 73.43 & 57.81 \\
UST-SSM \cite{ustssm} & ICCV'25 & TPC & 71.50 & 48.40 \\
\midrule
\rowcolor{gray!10}
\textbf{DAP-Net} & -- & MPC & \textbf{80.72} & \textbf{81.82} \\
\bottomrule
\end{tabular}
\end{table*}

\section{Experiment}

\subsection{Implementation Details}
Experiments are conducted on two NVIDIA RTX 4090 GPUs. DAP-Net is trained for 150 epochs (batch size 128, learning rate 0.01, weight decay $1\times10^{-4}$). We set $P_{\text{goal}}=1024$ for DGR and a 5-fold duplication for TMPD's Fast Branch. Using PointMLP~\cite{ma2022rethinking} as the backbone, TAM aligns features with a frozen CLIP~\cite{clip} text encoder via the template \textit{"a mmWave point cloud of a person [CLS]"}. Optimization uses standard cross-entropy loss.

\subsection{Compare with SOTA}

We evaluate representative methods on the UniMM-HAR C-Sub and C-Set in Table~\ref{tab:sota_comparison}, including mmWave radar point cloud methods and LiDAR-based dense point cloud methods (e.g., static point clouds and temporal point clouds). Existing dense point cloud methods fail on mmWave point clouds due to sparsity and noise, and homogeneous mmWave methods perform poorly under heterogeneous settings. In contrast, DAP-Net consistently outperforms all methods, demonstrating its effectiveness, heterogeneous cross-source generalizability.

\subsection{Ablation Study}
\begin{table}[t]
\centering
\begin{minipage}[t]{0.48\textwidth}
\centering
\small
\setlength{\tabcolsep}{0.7pt}
\caption{Impact of DAP-Net modules across backbones.}
\label{tab:backbone_dap}
\vspace{-0.1cm}
\resizebox{\textwidth}{!}{%
\begin{tabular}{l l c}
\toprule
\textbf{Backbone} & \textbf{Module} & \textbf{Acc (\%)} \\
\midrule
\multirow{3}{*}{PointMLP \cite{ma2022rethinking}} 
 & -- & 71.06 \\
 & +D$^2$R & $80.00^{\uparrow \textcolor{blue}{8.94}}$ \\
 & +D$^2$R+TAM & $80.72^{\uparrow \textcolor{blue}{9.66}}$ \\
\midrule
\multirow{3}{*}{UST-SSM \cite{ustssm}} 
 & -- & 71.50 \\
 & +D$^2$R & $74.00^{\uparrow \textcolor{blue}{2.50}}$ \\
 & +D$^2$R+TAM & $74.94^{\uparrow \textcolor{blue}{3.44}}$ \\
\midrule
\multirow{3}{*}{PST-Transformer\cite{pst_transformer}} 
 & -- & 63.13 \\
 & +D$^2$R & $76.73^{\uparrow \textcolor{blue}{13.60}}$ \\
 & +D$^2$R+TAM & $78.21^{\uparrow \textcolor{blue}{15.08}}$ \\
\bottomrule
\end{tabular}%
}
\end{minipage}
\hfill
\begin{minipage}[t]{0.48\textwidth}
\centering
\small
\caption{Impact of D$^2$R modules (DGR and MFR).}
\label{tab:d2r_ablation}
\vspace{-2mm}
\renewcommand{\arraystretch}{0.7}
\resizebox{\textwidth}{!}{%
\begin{tabular}{l cc c}
\toprule
\textbf{Backbone} & \multicolumn{2}{c}{\textbf{D$^2$R}} & \textbf{Acc (\%)} \\
\cmidrule(lr){2-3}
 & \textbf{DGR} & \textbf{MFR} &  \\
\midrule

\multirow{4}{*}{PointMLP \cite{ma2022rethinking}} 
 & \ding{55} & \ding{55} & 71.06 \\
 & \ding{51} & \ding{55} & $76.97^{\uparrow \textcolor{blue}{5.91}}$ \\
 & \ding{55} & \ding{51} & $78.76^{\uparrow \textcolor{blue}{7.70}}$ \\
 & \ding{51} & \ding{51} & $\mathbf{80.00}^{\uparrow \textcolor{blue}{8.94}}$ \\

\midrule

\multirow{3}{*}{UST-SSM \cite{ustssm}} 
 & \ding{55} & \ding{55} & 71.50 \\
 & \ding{51} & \ding{55} & $73.69^{\uparrow \textcolor{blue}{2.19}}$ \\
 & \ding{51} & \ding{51} & $\mathbf{74.00}^{\uparrow \textcolor{blue}{2.50}}$ \\

\midrule

\multirow{3}{*}{PST-Transformer \cite{pst_transformer}} 
 & \ding{55} & \ding{55} & 63.13 \\
 & \ding{51} & \ding{55} & $72.31^{\uparrow \textcolor{blue}{9.18}}$ \\
 & \ding{51} & \ding{51} & $\mathbf{76.73}^{\uparrow \textcolor{blue}{13.60}}$ \\

\bottomrule
\end{tabular}%
}
\end{minipage}

\end{table}

\begin{table}[t]
\centering

\begin{minipage}[t]{0.55\linewidth} 
\centering
\footnotesize
\caption{Impact of different point split strategies and fast branch densification in TMPD.}
\label{tab:tri_fast}
\vspace{-2mm} 
\resizebox{\linewidth}{!}{%
\begin{tabular}{l c c c}
\toprule
\textbf{Backbone} 
& \textbf{Points Split} 
& \makecell{\textbf{Fast Branch} \\ \textbf{Densification}} 
& \textbf{Acc (\%)} \\
\midrule
\multirow{4}{*}{PointMLP \cite{ma2022rethinking}} 
 & -- & -- & 71.06 \\
 & 0.2 quantile & MLP densification & 71.00$^{\downarrow \textcolor{blue}{0.06}}$ \\
 & 0.2 quantile & $r$-fold duplication & 76.46$^{\uparrow \textcolor{blue}{5.40}}$ \\
 & DSQ & $r$-fold duplication & 76.97$^{\uparrow \textcolor{blue}{5.91}}$ \\
\midrule
\multirow{4}{*}{UST-SSM \cite{ustssm}} 
 & -- & -- & 71.50 \\
 & 0.2 quantile & MLP densification & 71.40$^{\downarrow \textcolor{blue}{0.10}}$ \\
 & 0.2 quantile & $r$-fold duplication & 73.40$^{\uparrow \textcolor{blue}{1.90}}$ \\
 & DSQ & $r$-fold duplication & 74.00$^{\uparrow \textcolor{blue}{2.50}}$ \\
\bottomrule
\end{tabular}
}
\end{minipage}
\hfill
\begin{minipage}[t]{0.39\linewidth} 
\centering
\scriptsize
\renewcommand{\arraystretch}{1.2}
\setlength{\tabcolsep}{4pt}
\caption{Impact of different point split quantiles.}
\label{tab:quantile_comparison}
\resizebox{\linewidth}{!}{%
\begin{tabular}{l c c}
\toprule
\textbf{Quantile} & \makecell{\textbf{Split} \\ \textbf{Type}} & \textbf{Acc (\%)} \\
\midrule
--  & --      & 71.06 \\
0.2 & Fixed      & 76.46$^{\uparrow \textcolor{blue}{5.40}}$ \\
0.3 & Fixed      & 76.55$^{\uparrow \textcolor{blue}{5.40}}$ \\
0.5 & Fixed      & 76.14$^{\uparrow \textcolor{blue}{5.08}}$ \\
0.8 & Fixed      & 76.12$^{\uparrow \textcolor{blue}{5.06}}$ \\
\rowcolor{gray!10}
DSQ & Learnable  & 76.97$^{\uparrow \textcolor{blue}{5.91}}$ \\
\bottomrule
\end{tabular}%
}
\end{minipage}

\end{table}
We perform ablation studies on UniMM-HAR C-Sub to systematically investigate the contributions of each component in DAP-Net.

\noindent \textbf{Evaluation of Backbone Generalization}
As shown in Table~\ref{tab:backbone_dap}, integrating the modules of DAP-Net leads to substantial accuracy gains across all backbones. This improvement arises from D$^2$R’s intra-modal enhancement and TAM’s cross-modal alignment, which increase the backbone’s adaptability to mmWave point cloud characteristics and its robustness to heterogeneous distributions, thereby unlocking the full potential of dense point cloud backbones on mmWave data.

\noindent \textbf{Evaluation of D$^2$R}
In Table~\ref{tab:d2r_ablation}, we ablate D$^2$R on three representative backbones. Both DGR and MFR individually improve performance, and their combination achieves the best results, demonstrating that D$^2$R is backbone-agnostic and its modules provide complementary benefits.

\noindent \textbf{Evaluation of DSQ and TMPD}
We conduct ablation experiments to evaluate the individual and combined effects of DSQ and TMPD under different point splitting and densification strategies. As shown in Table~\ref{tab:tri_fast}, the proposed combination of DSQ and $r$-fold duplication of TMPD consistently achieves the best performance across different backbones, demonstrating the effectiveness of jointly modeling motion-aware point selection and densification.

\noindent \textbf{Evaluation of DSQ}
We compare the proposed learnable soft quantile in DSQ with fixed hard quantiles (0.2, 0.5, 0.8) and a variant without quantile-based splitting. As shown in Table~\ref{tab:quantile_comparison}, the adaptive soft quantile consistently achieves superior performance, demonstrating the benefit of learning a data-dependent threshold for motion-aware point selection.

\noindent \textbf{Evaluation of DGR}
In Table~\ref{tab:pointcloud_densification}, we compare DGR with several common point cloud densification strategies under a fixed number of output points. DGR achieves the best performance, substantially outperforming naive duplication and MLP-based generation, demonstrating that Doppler-guided geometric reparameterization is more effective for enhancing sparse mmWave representations.

\begin{table}[t]
\centering

\begin{minipage}[t]{0.32\linewidth}
\vspace{0pt} 
\centering
\caption{Impact of point cloud densification.}
\label{tab:pointcloud_densification}
\renewcommand{\arraystretch}{1.1}
\setlength{\tabcolsep}{4pt}
\vspace{-1mm}
\resizebox{\linewidth}{!}{
\begin{tabular}{l c}
\toprule
\textbf{Densification} & \textbf{Acc (\%)} \\
\midrule
Repeat Sampling & 71.06 \\ 
Super-frame Fusion & 59.72$^{\downarrow \textcolor{blue}{11.34}}$ \\
MLP & 43.54$^{\downarrow \textcolor{blue}{27.52}}$ \\
\rowcolor{gray!10}
DGR & 76.97$^{\uparrow \textcolor{blue}{5.91}}$ \\
\bottomrule
\end{tabular}%
}
\end{minipage}
\hfill
\begin{minipage}[t]{0.65\linewidth}
\vspace{0pt} 
\centering
\setlength{\tabcolsep}{4pt}
\renewcommand{\arraystretch}{1.3}
\caption{Impact of different text prompts and encoders.}
\label{tab:text_prompt}
\resizebox{\linewidth}{!}{
\begin{tabular}{l l c c}
\toprule
\textbf{Method} & \textbf{Action Text Prompt} & \textbf{Text Encoder} & \textbf{Acc (\%)} \\
\toprule
Baseline & -- & -- & 80.00 \\
\midrule
w/ TAM & [CLS] & CLIP \cite{clip} & 80.71$^{\uparrow \textcolor{blue}{0.71}}$ \\
w/ TAM & A person performing [CLS] & CLIP \cite{clip} & 80.65$^{\uparrow \textcolor{blue}{0.65}}$ \\
w/ TAM & a mmWave point cloud of a person [CLS] & SBERT \cite{reimers2019sentence} & 80.67$^{\uparrow \textcolor{blue}{0.67}}$ \\
w/ TAM & a mmWave point cloud of a person [CLS] & CLIP \cite{clip} & 80.72$^{\uparrow \textcolor{blue}{0.72}}$ \\
\bottomrule
\end{tabular}
}
\end{minipage}

\end{table}

\begin{figure*}[t]
\centering

\begin{minipage}[t]{0.48\linewidth}
\vspace{0pt} 
\centering
\includegraphics[width=0.8\linewidth]{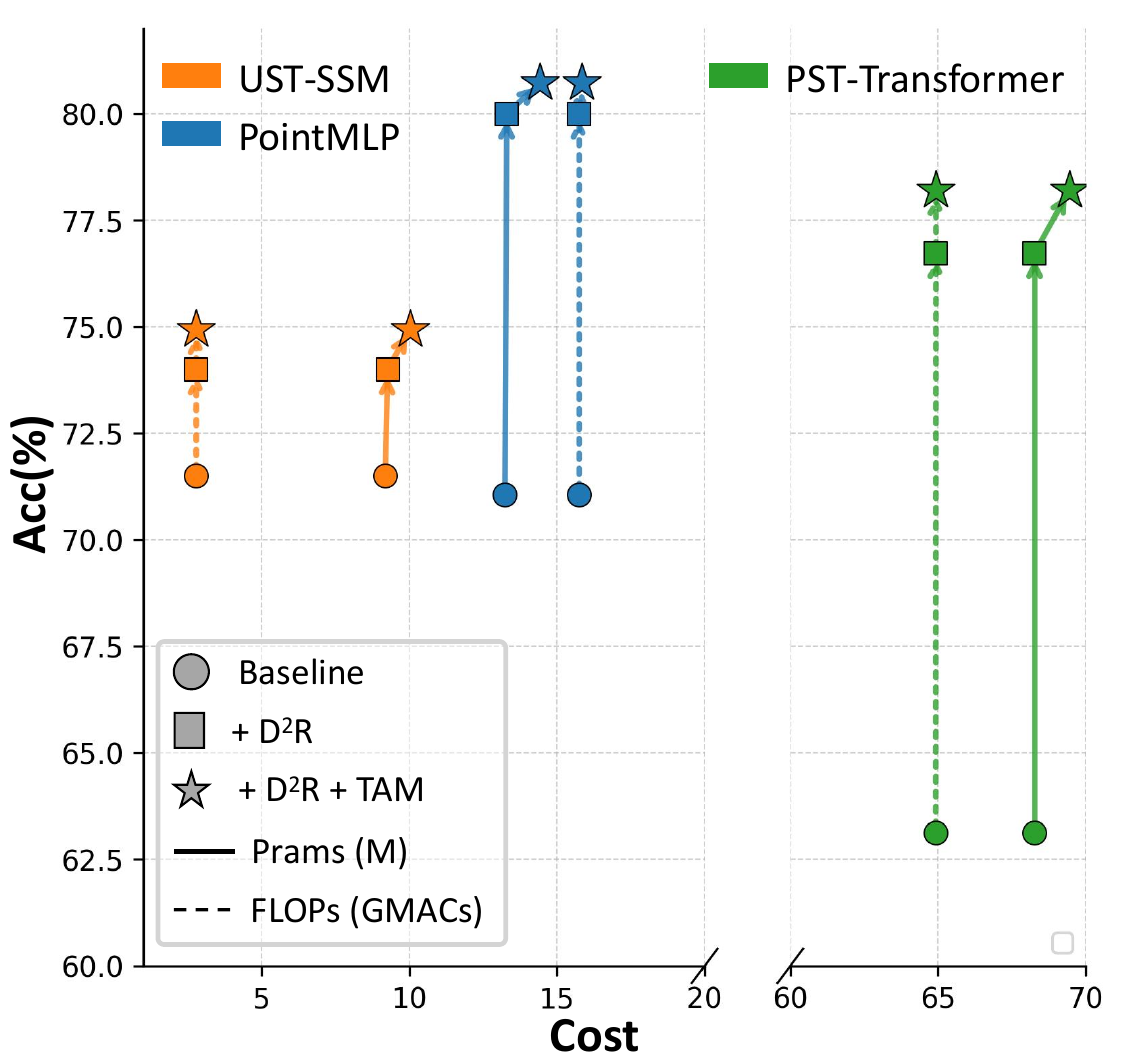}
\caption{Accuracy–Cost trade-off across different backbones.}
\label{fig:efficiency}
\end{minipage}
\hfill
\begin{minipage}[t]{0.48\linewidth}
\vspace{0pt} 
\centering
\includegraphics[width=0.8\linewidth]{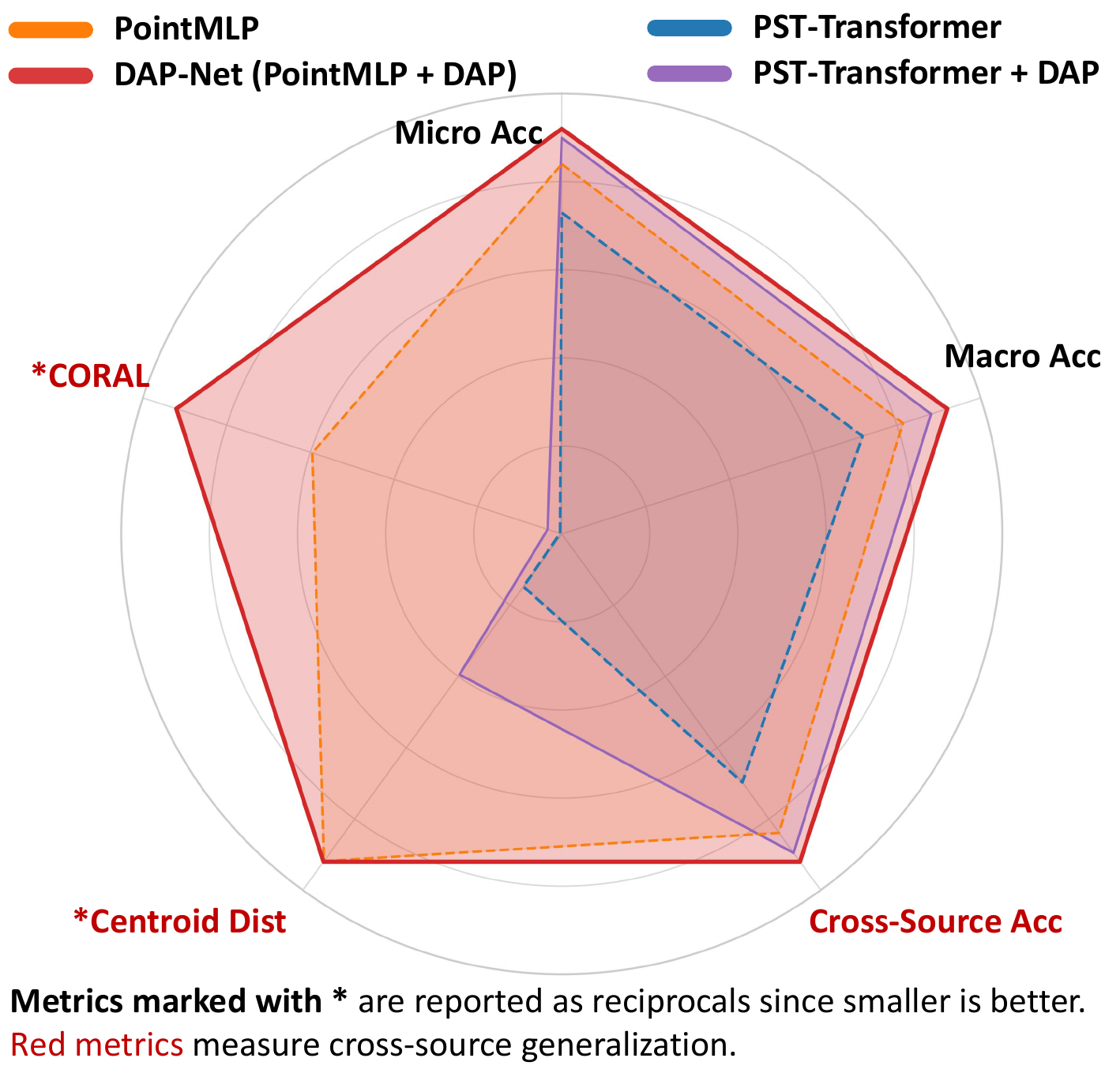}
\caption{Accuracy and cross-source generalization comparison.}
\label{fig:radar}
\end{minipage}

\end{figure*}

\noindent \textbf{Evaluation of TAM}
We evaluate TAM with various prompt templates and text encoders in Table~\ref{tab:text_prompt}. All combinations consistently yield performance gains, indicating that TAM consistently benefits from diverse text encoders and prompts, and can easily scale to more advanced text models such as LLMs.

\noindent \textbf{Evaluation of Efficiency}
We analyze the modules of DAP-Net (D²R and TAM) in terms of accuracy–cost trade-offs across different baselines in Fig.~\ref{fig:efficiency}. Both modules improve performance while incurring negligible parameter and FLOPs overhead, with D²R contributing particularly significant gains.

\noindent \textbf{Evaluation of Cross-Source Performance}
Fig.~\ref{fig:radar} shows performance on heterogeneous multi-source data, where \textit{Cross-Source Acc} measures action consistency across sources, \textit{Centroid Distance} quantifies shifts between source feature centroids, and \textit{CORAL} evaluates covariance differences across sources (see Supplementary for metric details). DAP-Net achieves the highest overall accuracy while excelling in cross-source metrics. Compared with the baseline, adding DAP-Net modules significantly improves heterogeneous cross-source metrics.

\section{Conclusion}
We construct the UniMM-HAR dataset and propose DAP-Net for heterogeneous multi-source mmWave point clouds. UniMM-HAR is the first and largest mmWave point cloud HAR benchmark designed for heterogeneous multi-source scenarios, standardizing three distinct radar configurations and providing C-Sub and C-Set evaluation protocols to rigorously assess cross-source generalization. To mitigate source-specific biases, DAP-Net leverages Doppler as a motion prior, performing sample-adaptive geometric densification and feature recalibration via D²R, and aligns features with a pretrained textual space through TAM to provide stable semantic anchors. Extensive experiments demonstrate that DAP-Net significantly improves both recognition accuracy and cross-source robustness.

\section*{Acknowledgements}
This work was supported by National Natural Science Foundation of China (No. 62473007), Guangdong Outstanding Youth Fund (No. 2026B1515020015), Shenzhen Innovation in Science and Technology Foundation for The Excellent Youth Scholars (No. RCYX20231211090248064).

%
%
\bibliographystyle{splncs04}
\bibliography{main}

@String(PAMI  = {IEEE TPAMI})

@String(CVPR  = {CVPR})

@String(ICCV  = {ICCV})

@String(ECCV  = {ECCV})

@String(NeurIPS = {NeurIPS})

@String(ICML  = {ICML})

@String(ICLR  = {ICLR})

@String(AAAI  = {AAAI})

@String(ICASSP=	{ICASSP})

@String(TMM   =	{IEEE TMM})

@String(ACMMM = {ACM MM})

@String(PR    = {PR})

@String(PAMI  = {IEEE Trans. Pattern Anal. Mach. Intell.})

@String(CVPR  = {IEEE Conf. Comput. Vis. Pattern Recog.})

@String(ICCV  = {Int. Conf. Comput. Vis.})

@String(ECCV  = {Eur. Conf. Comput. Vis.})

@String(NeurIPS = {Adv. Neural Inform. Process. Syst.})

@String(ICML  = {Int. Conf. Mach. Learn.})

@String(ICLR  = {Int. Conf. Learn. Represent.})

@String(TMM   = {IEEE Trans. Multimedia})

@String(ACMMM = {ACM Int. Conf. Multimedia})

@String(PR    = {Pattern Recognition})

@String(CIKM = {ACM Int. Conf. Inf. Knowl. Manag.})

@String(IMWUT = {ACM Interact. Mob. Wearable Ubiquitous Technol.})

@String(TMC    = {IEEE Trans. Mobile Comput.})

@String(EMBC = {IEEE Eng. Med. Biol. Soc. Conf.})

@String(VTC     = {IEEE Veh. Technol. Conf.})

@String(EMNLP = {Conf. Empir. Methods Nat. Lang. Process.})

@String(ICRA = {IEEE Int. Conf. Robot. Autom.})

@String(ICMEW = {IEEE Int. Conf. Multimedia Expo Workshops})

@String(WACV = {IEEE/CVF Winter Conf. Appl. Comput. Vis.})

@inproceedings{singh2019radhar,
  title={Radhar: Human activity recognition from point clouds generated through a millimeter-wave radar},
  author={Singh, Akash Deep and Sandha, Sandeep Singh and Garcia, Luis and Srivastava, Mani},
  booktitle={ACM Workshop Millimeter-wave Netw. Sens. Syst.},
  pages={51--56},
  year={2019}
}

@inproceedings{shao2024fast,
  title={Fast Human Action Recognition via Millimeter Wave Radar Point Cloud Sequences Learning},
  author={Shao, Tongfei and Du, Zheyu and Li, Chuanyou and Wu, Tianxing and Wang, Meng},
  booktitle=CIKM,
  pages={2024--2033},
  year={2024}
}

@inproceedings{qi2017pointnet,
  title={Pointnet: Deep learning on point sets for 3d classification and segmentation},
  author={Qi, Charles R and Su, Hao and Mo, Kaichun and Guibas, Leonidas J},
  booktitle=CVPR,
  pages={652--660},
  year={2017}
}

@article{phan2018dgcnn,
  title={Dgcnn: A convolutional neural network over large-scale labeled graphs},
  author={Phan, Anh Viet and Le Nguyen, Minh and Nguyen, Yen Lam Hoang and Bui, Lam Thu},
  journal={Neural Networks},
  volume={108},
  pages={533--543},
  year={2018},
  publisher={Elsevier}
}

@inproceedings{ma2022rethinking,
  title={Rethinking Network Design and Local Geometry in Point Cloud: A Simple Residual MLP Framework},
  author={Ma, Xu and Qin, Can and You, Haoxuan and Ran, Haoxi and Fu, Yun},
  booktitle=ICLR,
  year={2022}
}

@inproceedings{huang2023clip2point,
  title={Clip2point: Transfer clip to point cloud classification with image-depth pre-training},
  author={Huang, Tianyu and Dong, Bowen and Yang, Yunhan and Huang, Xiaoshui and Lau, Rynson WH and Ouyang, Wanli and Zuo, Wangmeng},
  booktitle=ICCV,
  pages={22157--22167},
  year={2023}
}

@article{courbariaux2016binarized,
  title={Binarized neural networks: Training deep neural networks with weights and activations constrained to+ 1 or-1},
  author={Courbariaux, Matthieu and Hubara, Itay and Soudry, Daniel and El-Yaniv, Ran and Bengio, Yoshua},
  journal={arXiv preprint arXiv:1602.02830},
  year={2016}
}

@article{bengio2013estimating,
  title={Estimating or propagating gradients through stochastic neurons for conditional computation},
  author={Bengio, Yoshua and L{\'e}onard, Nicholas and Courville, Aaron},
  journal={arXiv preprint arXiv:1308.3432},
  year={2013}
}

@article{an2021mars,
  title={Mars: mmwave-based assistive rehabilitation system for smart healthcare},
  author={An, Sizhe and Ogras, Umit Y},
  journal={ACM Trans. Embed. Comput. Syst.},
  volume={20},
  number={5s},
  pages={1--22},
  year={2021},
  publisher={ACM New York, NY}
}

@inproceedings{m_activity,
  title={m-activity: Accurate and real-time human activity recognition via millimeter wave radar},
  author={Wang, Yuheng and Liu, Haipeng and Cui, Kening and Zhou, Anfu and Li, Wensheng and Ma, Huadong},
  booktitle=ICASSP,
  pages={8298--8302},
  year={2021},
  organization={IEEE}
}

@article{mRI,
  title={mri: Multi-modal 3d human pose estimation dataset using mmwave, rgb-d, and inertial sensors},
  author={An, Sizhe and Li, Yin and Ogras, Umit},
  journal=NeurIPS,
  volume={35},
  pages={27414--27426},
  year={2022}
}

@article{cui2023milipoint,
  title={Milipoint: A point cloud dataset for mmwave radar},
  author={Cui, Han and Zhong, Shu and Wu, Jiacheng and Shen, Zichao and Dahnoun, Naim and Zhao, Yiren},
  journal=NeurIPS,
  volume={36},
  pages={62713--62726},
  year={2023}
}

@article{mmfi,
  title={Mm-fi: Multi-modal non-intrusive 4d human dataset for versatile wireless sensing},
  author={Yang, Jianfei and Huang, He and Zhou, Yunjiao and Chen, Xinyan and Xu, Yuecong and Yuan, Shenghai and Zou, Han and Lu, Chris Xiaoxuan and Xie, Lihua},
  journal=NeurIPS,
  volume={36},
  pages={18756--18768},
  year={2023}
}

@article{hor2023mvdoppler,
  title={Mvdoppler: Unleashing the power of multi-view doppler for micromotion-based gait classification},
  author={Hor, Soheil and Yang, Shubo and Choi, Jaeho and Arbabian, Amin},
  journal=NeurIPS,
  volume={36},
  pages={58064--58074},
  year={2023}
}

@article{mmparse,
  title={Human parsing with joint learning for dynamic mmwave radar point cloud},
  author={Wang, Shuai and Cao, Dongjiang and Liu, Ruofeng and Jiang, Wenchao and Yao, Tianshun and Lu, Chris Xiaoxuan},
  journal=IMWUT,
  volume={7},
  number={1},
  pages={1--22},
  year={2023},
  publisher={ACM New York, NY, USA}
}

@inproceedings{fastHAR,
  title={Fast Human Action Recognition via Millimeter Wave Radar Point Cloud Sequences Learning},
  author={Shao, Tongfei and Du, Zheyu and Li, Chuanyou and Wu, Tianxing and Wang, Meng},
  booktitle=CIKM,
  pages={2024--2033},
  year={2024}
}

@inproceedings{ding2024milliflow,
  title={milliflow: Scene flow estimation on mmwave radar point cloud for human motion sensing},
  author={Ding, Fangqiang and Luo, Zhen and Zhao, Peijun and Lu, Chris Xiaoxuan},
  booktitle=ECCV,
  pages={202--221},
  year={2024},
  organization={Springer}
}

@inproceedings{choi2025mvdopplerpose,
  title={MVDoppler-Pose: Multi-Modal Multi-View mmWave Sensing for Long-Distance Self-Occluded Human Walking Pose Estimation},
  author={Choi, Jaeho and Hor, Soheil and Yang, Shubo and Arbabian, Amin},
  booktitle=CVPR,
  pages={27750--27759},
  year={2025}
}

@article{zhou2025mmmulti,
  title={mmMulti: Multi-person Action Recognition Based on Multi-task Learning Using Millimeter Waves},
  author={Zhou, Rui and Li, Songlin and Zhang, Hongwang and Liu, Chenxu and Sun, Jiajun},
  journal=IMWUT,
  volume={9},
  number={2},
  pages={1--25},
  year={2025},
  publisher={ACM New York, NY, USA}
}

@article{zhang2018cnn,
  title={Real-time human motion behavior detection via CNN using mmWave radar},
  author={Zhang, Renyuan and Cao, Siyang},
  journal={IEEE Sensors Letters},
  volume={3},
  number={2},
  pages={1--4},
  year={2018},
  publisher={IEEE}
}

@inproceedings{xue2021mmmesh,
  title={mmMesh: Towards 3D real-time dynamic human mesh construction using millimeter-wave},
  author={Xue, Hongfei and Ju, Yan and Miao, Chenglin and Wang, Yijiang and Wang, Shiyang and Zhang, Aidong and Su, Lu},
  booktitle={ACM MobiSys},
  pages={269--282},
  year={2021}
}

@inproceedings{mmgait,
  title={Gait recognition for co-existing multiple people using millimeter wave sensing},
  author={Meng, Zhen and Fu, Song and Yan, Jie and Liang, Hongyuan and Zhou, Anfu and Zhu, Shilin and Ma, Huadong and Liu, Jianhua and Yang, Ning},
  booktitle=AAAI,
  volume={34},
  pages={849--856},
  year={2020}
}

@inproceedings{feichtenhofer2019slowfast,
  title={Slowfast networks for video recognition},
  author={Feichtenhofer, Christoph and Fan, Haoqi and Malik, Jitendra and He, Kaiming},
  booktitle=ICCV,
  pages={6202--6211},
  year={2019}
}

@article{liu2017enhanced,
  title={Enhanced skeleton visualization for view invariant human action recognition},
  author={Liu, Mengyuan and Liu, Hong and Chen, Chen},
  journal=PR,
  volume={68},
  pages={346--362},
  year={2017},
  publisher={Elsevier}
}

@inproceedings{duan2022revisiting,
  title={Revisiting skeleton-based action recognition},
  author={Duan, Haodong and Zhao, Yue and Chen, Kai and Lin, Dahua and Dai, Bo},
  booktitle=CVPR,
  pages={2969--2978},
  year={2022}
}

@article{bruce2022mmnet,
  title={Mmnet: A model-based multimodal network for human action recognition in rgb-d videos},
  author={Bruce, XB and Liu, Yan and Zhang, Xiang and Zhong, Sheng-hua and Chan, Keith CC},
  journal=PAMI,
  volume={45},
  number={3},
  pages={3522--3538},
  year={2022},
  publisher={IEEE}
}

@article{xia2024timestamp,
  title={Timestamp-supervised wearable-based activity segmentation and recognition with contrastive learning and order-preserving optimal transport},
  author={Xia, Songpengcheng and Chu, Lei and Pei, Ling and Yang, Jiarui and Yu, Wenxian and Qiu, Robert C},
  journal=TMC,
  volume={23},
  number={12},
  pages={10734--10751},
  year={2024},
  publisher={IEEE}
}

@inproceedings{xu2025ai,
  title={AI-driven personalized fall prevention for older adults},
  author={Xu, Katherine},
  booktitle=AAAI,
  volume={39},
  number={28},
  pages={29610--29612},
  year={2025}
}

@inproceedings{yan2018spatial,
  author = {Yan, Sijie and Xiong, Yuanjun and Lin, Dahua},
  title = {Spatial temporal graph convolutional networks for skeleton-based action recognition},
  booktitle = AAAI,
  year = {2018},
  volume={32},
  number={1},
}

@InProceedings{Chen_2021_ICCV,
    author = {Chen, Yuxin and Zhang, Ziqi and Yuan, Chunfeng and Li, Bing and Deng, Ying and Hu, Weiming},
    title = {Channel-Wise Topology Refinement Graph Convolution for Skeleton-Based Action Recognition},
    booktitle = ICCV,
    year = {2021},
    pages = {13359-13368},
}

@ARTICLE{10113233,
  author = {Liu, Jinfu and Wang, Xinshun and Wang, Can and Gao, Yuan and Liu, Mengyuan},
  title = {Temporal Decoupling Graph Convolutional Network for Skeleton-based Gesture Recognition}, 
  journal = TMM, 
  year = {2024},
  volume={26},
  pages={811--823},
}

@article{palipana2021pantomime,
  title={Pantomime: Mid-air gesture recognition with sparse millimeter-wave radar point clouds},
  author={Palipana, Sameera and Salami, Dariush and Leiva, Luis A and Sigg, Stephan},
  journal=IMWUT,
  volume={5},
  number={1},
  pages={1--27},
  year={2021},
  publisher={ACM New York, NY, USA}
}

@article{salami2022tesla,
  title={Tesla-rapture: A lightweight gesture recognition system from mmwave radar sparse point clouds},
  author={Salami, Dariush and Hasibi, Ramin and Palipana, Sameera and Popovski, Petar and Michoel, Tom and Sigg, Stephan},
  journal=TMC,
  volume={22},
  number={8},
  pages={4946--4960},
  year={2022},
  publisher={IEEE}
}

@inproceedings{wan2014gesture,
  title={Gesture recognition for smart home applications using portable radar sensors},
  author={Wan, Qian and Li, Yiran and Li, Changzhi and Pal, Ranadip},
  booktitle=EMBC,
  pages={6414--6417},
  year={2014},
  organization={IEEE}
}

@inproceedings{zhang2022pointclip,
  title={Pointclip: Point cloud understanding by clip},
  author={Zhang, Renrui and Guo, Ziyu and Zhang, Wei and Li, Kunchang and Miao, Xupeng and Cui, Bin and Qiao, Yu and Gao, Peng and Li, Hongsheng},
  booktitle=CVPR,
  pages={8552--8562},
  year={2022}
}

@inproceedings{perez2018film,
  title={Film: Visual reasoning with a general conditioning layer},
  author={Perez, Ethan and Strub, Florian and De Vries, Harm and Dumoulin, Vincent and Courville, Aaron},
  booktitle=AAAI,
  volume={32},
  number={1},
  year={2018}
}

@INPROCEEDINGS{ra_2,
  author={Yu, Jih-Tsun and Yen, Li and Tseng, Po-Hsuan},
  booktitle=VTC, 
  title={mmWave Radar-based Hand Gesture Recognition using Range-Angle Image}, 
  year={2020},
  pages={1-5}
}

@article{sengupta2022mmpose,
  title={mmpose-nlp: A natural language processing approach to precise skeletal pose estimation using mmwave radars},
  author={Sengupta, Arindam and Cao, Siyang},
  journal={IEEE Trans. Neural Netw. Learn. Syst.},
  volume={34},
  number={11},
  pages={8418--8429},
  year={2022},
  publisher={IEEE}
}

@inproceedings{lee2023hupr,
  title={Hupr: A benchmark for human pose estimation using millimeter wave radar},
  author={Lee, Shih-Po and Kini, Niraj Prakash and Peng, Wen-Hsiao and Ma, Ching-Wen and Hwang, Jenq-Neng},
  booktitle=WACV,
  pages={5715--5724},
  year={2023}
}

@article{zhao2023cubelearn,
  title={Cubelearn: End-to-end learning for human motion recognition from raw mmwave radar signals},
  author={Zhao, Peijun and Lu, Chris Xiaoxuan and Wang, Bing and Trigoni, Niki and Markham, Andrew},
  journal={IEEE Internet of Things Journal},
  volume={10},
  number={12},
  pages={10236--10249},
  year={2023},
  publisher={IEEE}
}

@inproceedings{mahmood2019amass,
  title={AMASS: Archive of motion capture as surface shapes},
  author={Mahmood, Naureen and Ghorbani, Nima and Troje, Nikolaus F and Pons-Moll, Gerard and Black, Michael J},
  booktitle=ICCV,
  pages={5442--5451},
  year={2019}
}

@article{pst_transformer,
  title={Point spatio-temporal transformer networks for point cloud video modeling},
  author={Fan, Hehe and Yang, Yi and Kankanhalli, Mohan},
  journal=PAMI,
  volume={45},
  number={2},
  pages={2181--2192},
  year={2022},
  publisher={IEEE}
}

@inproceedings{kim2021point,
  title={Point cloud augmentation with weighted local transformations},
  author={Kim, Sihyeon and Lee, Sanghyeok and Hwang, Dasol and Lee, Jaewon and Hwang, Seong Jae and Kim, Hyunwoo J},
  booktitle=ICCV,
  pages={548--557},
  year={2021}
}

@inproceedings{ben20243dinaction,
  title={3dinaction: Understanding human actions in 3d point clouds},
  author={Ben-Shabat, Yizhak and Shrout, Oren and Gould, Stephen},
  booktitle=CVPR,
  pages={19978--19987},
  year={2024}
}

@inproceedings{ustssm,
  title={UST-SSM: Unified Spatio-Temporal State Space Models for Point Cloud Video Modeling},
  author={Li, Peiming and Wang, Ziyi and Yuan, Yulin and Liu, Hong and Meng, Xiangming and Yuan, Junsong and Liu, Mengyuan},
  booktitle=ICCV,
  pages={6738--6747},
  year={2025}
}

@inproceedings{liu2025mamba4d,
  title={Mamba4d: Efficient 4d point cloud video understanding with disentangled spatial-temporal state space models},
  author={Liu, Jiuming and Han, Jinru and Liu, Lihao and Aviles-Rivero, Angelica I and Jiang, Chaokang and Liu, Zhe and Wang, Hesheng},
  booktitle=CVPR,
  pages={17626--17636},
  year={2025}
}

@inproceedings{clip,
  title={Learning transferable visual models from natural language supervision},
  author={Radford, Alec and Kim, Jong Wook and Hallacy, Chris and Ramesh, Aditya and Goh, Gabriel and Agarwal, Sandhini and Sastry, Girish and Askell, Amanda and Mishkin, Pamela and Clark, Jack and others},
  booktitle=ICML,
  pages={8748--8763},
  year={2021},
  organization={PmLR}
}

@inproceedings{reimers2019sentence,
  title={Sentence-bert: Sentence embeddings using siamese bert-networks},
  author={Reimers, Nils and Gurevych, Iryna},
  booktitle=EMNLP,
  pages={3982--3992},
  year={2019}
}

@inproceedings{wang2016interacting,
  title={Interacting with soli: Exploring fine-grained dynamic gesture recognition in the radio-frequency spectrum},
  author={Wang, Saiwen and Song, Jie and Lien, Jaime and Poupyrev, Ivan and Hilliges, Otmar},
  booktitle={UIST},
  pages={851--860},
  year={2016}
}

@article{fhager2019pulsed,
  title={Pulsed millimeter wave radar for hand gesture sensing and classification},
  author={Fhager, Lars Ohlsson and Heunisch, Sebastian and Dahlberg, Hannes and Evertsson, Anton and Wernersson, Lars-Erik},
  journal={IEEE Sensors Letters},
  volume={3},
  number={12},
  pages={1--4},
  year={2019},
  publisher={IEEE}
}

@inproceedings{fan2026m4human,
  title={M4human: A large-scale multimodal mmwave radar benchmark for human mesh reconstruction},
  author={Fan, Junqiao and Zhou, Yunjiao and Yang, Yizhuo and Cui, Xinyuan and Zhang, Jiarui and Xie, Lihua and Yang, Jianfei and Lu, Chris Xiaoxuan and Ding, Fangqiang},
  booktitle=CVPR,
  pages={42836--42846},
  year={2026}
}

@ARTICLE{microd_1,
  author={Biswas, Sabyasachi and Manavi Alam, Ahmed and Gurbuz, Ali C.},
  journal={IEEE Transactions on Radar Systems}, 
  title={HRSpecNET: A Deep Learning-Based High-Resolution Radar Micro-Doppler Signature Reconstruction for Improved HAR Classification}, 
  year={2024},
  volume={2},
  pages={484-497},
}

@Article{microd_2,
AUTHOR = {Tan, Tan-Hsu and Tian, Jia-Hong and Sharma, Alok Kumar and Liu, Shing-Hong and Huang, Yung-Fa},
TITLE = {Human Activity Recognition Based on Deep Learning and Micro-Doppler Radar Data},
JOURNAL = {Sensors},
VOLUME = {24},
YEAR = {2024},
NUMBER = {8},
}

@article{microd_3,
  title={RadMamba: Efficient Human Activity Recognition Through a Radar-Based Micro-Doppler-Oriented Mamba State-Space Model},
  author={Wu, Yizhuo and Fioranelli, Francesco and Gao, Chang},
  journal={IEEE Transactions on Radar Systems},
  volume={4},
  pages={261--272},
  year={2025},
  publisher={IEEE}
}

@inproceedings{vox_1,
  title={Human Activity Recognition Based on 4D Millimeter-Wave Radar},
  author={Gao, Guixing and Liu, Quanli and Wang, Wei and Yu, Zichen and Liu, Xin},
  booktitle={Youth Academic Annual Conf. Chinese Assoc. Autom.},
  pages={1822--1828},
  year={2025},
  organization={IEEE}
}

@article{vox_2,
  title={Noninvasive human activity recognition using millimeter-wave radar},
  author={Yu, Chengxi and Xu, Zhezhuang and Yan, Kun and Chien, Ying-Ren and Fang, Shih-Hau and Wu, Hsiao-Chun},
  journal={IEEE Systems Journal},
  volume={16},
  number={2},
  pages={3036--3047},
  year={2022},
  publisher={IEEE}
}

@inproceedings{vox_3,
  title={Multi-har: Human activity recognition in multi-person scenes based on mmwave sensing},
  author={Zeng, Xianlin and Shi, Yiming and Zhou, Anfu},
  booktitle={IEEE Int. Conf. Comput. Commun.},
  pages={1789--1793},
  year={2022},
  organization={IEEE}
}

@article{luo2023edgeactnet,
  title={EdgeActNet: Edge intelligence-enabled human activity recognition using radar point cloud},
  author={Luo, Fei and Khan, Salabat and Li, Anna and Huang, Yandao and Wu, Kaishun},
  journal={IEEE Trans. Mobile Comput.},
  volume={23},
  number={5},
  pages={5479--5493},
  year={2023},
  publisher={IEEE}
}

@article{mmw_1,
  title={Millimeter wave radar-based human activity recognition for healthcare monitoring robot},
  author={Gu, Zhanzhong and He, Xiangjian and Fang, Gengfa and Xu, Chengpei and Xia, Feng and Jia, Wenjing},
  journal={arXiv preprint arXiv:2405.01882},
  year={2024}
}

@inproceedings{mmw_2,
  title={Enhanced Sparse Point Cloud Data Processing for Privacy-Aware Human Action Recognition},
  author={Tunau, Maimunatu and Zakka, Vincent Gbouna and Dai, Zhuangzhuang},
  booktitle={Int. Conf. AI Healthcare},
  pages={142--155},
  year={2025},
  organization={Springer}
}

@article{mmw_3,
  title={Multi-person action recognition based on millimeter-wave radar point cloud},
  author={Dang, Xiaochao and Fan, Kai and Li, Fenfang and Tang, Yangyang and Gao, Yifei and Wang, Yue},
  journal={Applied Sciences},
  volume={14},
  number={16},
  pages={7253},
  year={2024},
  publisher={MDPI}
}

@article{rd_1,
  title={OG-PCL: Efficient Sparse Point Cloud Processing for Human Activity Recognition},
  author={Yan, Jiuqi and Xu, Chendong and Liu, Dongyu},
  journal={arXiv preprint arXiv:2511.08910},
  year={2025}
}

@article{rd_2,
  title={Diffusion-based mmWave radar point cloud enhancement driven by range images},
  author={Wu, Ruixin and Li, Zihan and Wang, Jin and Xu, Xiangyu and Zheng, Zhi and Huang, Kaixiang and Lu, Guodong},
  journal={arXiv preprint arXiv:2503.02300},
  year={2025}
}

@ARTICLE{rd_3,
  author={Seo, Hong-Il and Bae, Ju-Won and Seo, Dong-Hoan},
  journal={IEEE Sensors Journal}, 
  title={Radar-Based Human Activity Recognition Using Adaptive Range Selection and Deep Neural Network}, 
  year={2026},
  pages={1-1}
}

@inproceedings{zhao2018rf,
  title={RF-based 3D skeletons},
  author={Zhao, Mingmin and Tian, Yonglong and Zhao, Hang and Alsheikh, Mohammad Abu and Li, Tianhong and Hristov, Rumen and Kabelac, Zachary and Katabi, Dina and Torralba, Antonio},
  booktitle={ACM SIGCOMM},
  pages={267--281},
  year={2018}
}

@inproceedings{torralba2011unbiased,
  title={Unbiased look at dataset bias},
  author={Torralba, Antonio and Efros, Alexei A},
  booktitle=CVPR,
  pages={1521--1528},
  year={2011},
  organization={IEEE}
}

@article{zhou2022domain,
  title={Domain generalization: A survey},
  author={Zhou, Kaiyang and Liu, Ziwei and Qiao, Yu and Xiang, Tao and Loy, Chen Change},
  journal=PAMI,
  volume={45},
  number={4},
  pages={4396--4415},
  year={2022},
  publisher={IEEE}
}

@inproceedings{liu2024unbiased,
  title={Unbiased faster r-cnn for single-source domain generalized object detection},
  author={Liu, Yajing and Zhou, Shijun and Liu, Xiyao and Hao, Chunhui and Fan, Baojie and Tian, Jiandong},
  booktitle=CVPR,
  pages={28838--28847},
  year={2024}
}

@inproceedings{chae2025doppler,
  title={Doppler-Aware LiDAR-RADAR Fusion for Weather-Robust 3D Detection},
  author={Chae, Yujeong and Park, Heejun and Kim, Hyeonseong and Yoon, Kuk-Jin},
  booktitle=ICCV,
  pages={27197--27208},
  year={2025}
}

@inproceedings{haitman2025doppdrive,
  title={DoppDrive: Doppler-Driven Temporal Aggregation for Improved Radar Object Detection},
  author={Haitman, Yuval and Bialer, Oded},
  booktitle=ICCV,
  pages={26085--26094},
  year={2025}
}

@inproceedings{prabhakara2023high,
  title={High resolution point clouds from mmwave radar},
  author={Prabhakara, Akarsh and Jin, Tao and Das, Arnav and Bhatt, Gantavya and Kumari, Lilly and Soltanaghai, Elahe and Bilmes, Jeff and Kumar, Swarun and Rowe, Anthony},
  booktitle=ICRA,
  pages={4135--4142},
  year={2023},
  organization={IEEE}
}

@article{liu2023echoes,
  title={Echoes beyond points: Unleashing the power of raw radar data in multi-modality fusion},
  author={Liu, Yang and Wang, Feng and Wang, Naiyan and Zhang, Zhao-Xiang},
  journal=NeurIPS,
  volume={36},
  pages={53964--53982},
  year={2023}
}

@ARTICLE{11278750,
  author={Liu, Mengyuan and Liu, Jinfu and Jiang, Yongkang and He, Bin},
  journal=PAMI, 
  title={Heatmap Pooling Network for Action Recognition From RGB Videos}, 
  year={2026},
  volume={48},
  number={3},
  pages={3726-3743}
}

@article{qi2017pointnet++,
  title={Pointnet++: Deep hierarchical feature learning on point sets in a metric space},
  author={Qi, Charles Ruizhongtai and Yi, Li and Su, Hao and Guibas, Leonidas J},
  journal=NeurIPS,
  volume={30},
  year={2017}
}

@inproceedings{liu2025revealing,
  title={Revealing key details to see differences: A novel prototypical perspective for skeleton-based action recognition},
  author={Liu, Hongda and Liu, Yunfan and Ren, Min and Wang, Hao and Wang, Yunlong and Sun, Zhenan},
  booktitle=CVPR,
  pages={29248--29257},
  year={2025}
}

@inproceedings{xu2024pointllm,
  title={Pointllm: Empowering large language models to understand point clouds},
  author={Xu, Runsen and Wang, Xiaolong and Wang, Tai and Chen, Yilun and Pang, Jiangmiao and Lin, Dahua},
  booktitle=ECCV,
  pages={131--147},
  year={2024},
  organization={Springer}
}

@inproceedings{hinojosa2022privhar,
  title={Privhar: Recognizing human actions from privacy-preserving lens},
  author={Hinojosa, Carlos and Marquez, Miguel and Arguello, Henry and Adeli, Ehsan and Fei-Fei, Li and Niebles, Juan Carlos},
  booktitle=ECCV,
  pages={314--332},
  year={2022},
  organization={Springer}
}

@inproceedings{deng2024vg4d,
  title={Vg4d: Vision-language model goes 4d video recognition},
  author={Deng, Zhichao and Li, Xiangtai and Li, Xia and Tong, Yunhai and Zhao, Shen and Liu, Mengyuan},
  booktitle=ICRA,
  pages={5014--5020},
  year={2024},
  organization={IEEE}
}

@misc{ti_iwr1443boost,
  author       = {{Texas Instruments}},
  title        = {{IWR1443BOOST} Single-Chip 77- and 79-GHz mmWave Sensor Evaluation Module},
  year         = {2014},
  howpublished = {\url{https://www.ti.com/tool/IWR1443BOOST}},
  note         = {Accessed: 2020-09-29}
}

@misc{ti_iwr6843,
  author       = {{Texas Instruments}},
  title        = {{IWR6843} Single-Chip 60-GHz mmWave Radar Sensor},
  year         = {2019},
  howpublished = {\url{https://www.ti.com/product/IWR6843}},
  note         = {Accessed: 2020-09-29}
}

@inproceedings{chen2022mmbody,
  title={mmbody benchmark: 3d body reconstruction dataset and analysis for millimeter wave radar},
  author={Chen, Anjun and Wang, Xiangyu and Zhu, Shaohao and Li, Yanxu and Chen, Jiming and Ye, Qi},
  booktitle=ACMMM,
  pages={3501--3510},
  year={2022}
}

@inproceedings{lai2026radarllm,
  title={Radarllm: empowering large language models to understand human motion from millimeter-wave point cloud sequence},
  author={Lai, Zengyuan and Yang, Jiarui and Xia, Songpengcheng and Lin, Lizhou and Sun, Lan and Wang, Renwen and Liu, Jianran and Wu, Qi and Pei, Ling},
  booktitle=AAAI,
  volume={40},
  number={7},
  pages={5791--5799},
  year={2026}
}

@inproceedings{guo2023point,
  title={Point transformer-based human activity recognition using high-dimensional radar point clouds},
  author={Guo, Zhongyuan and Guendel, Ronny G and Yarovoy, Alexander and Fioranelli, Francesco},
  booktitle={IEEE Radar Conference},
  pages={1--6},
  year={2023},
  organization={IEEE}
}

@inproceedings{liu2024multi,
  title={Multi-modality co-learning for efficient skeleton-based action recognition},
  author={Liu, Jinfu and Chen, Chen and Liu, Mengyuan},
  booktitle=ACMMM,
  pages={4909--4918},
  year={2024}
}

@inproceedings{liu2024hdbn,
  title={Hdbn: A novel hybrid dual-branch network for robust skeleton-based action recognition},
  author={Liu, Jinfu and Yin, Baiqiao and Lin, Jiaying and Wen, Jiajun and Li, Yue and Liu, Mengyuan},
  booktitle=ICMEW,
  pages={1--6},
  year={2024},
  organization={IEEE}
}

@inproceedings{lin2024sfmvit,
  title={SFMViT: SlowFast Meet ViT in Chaotic World},
  author={Lin, Jiaying and Wen, Jiajun and Liu, Mengyuan and Yue, L and Liu, Jinfu and Yin, Baiqiao},
  booktitle=ICMEW,
  pages={1--6},
  year={2024},
  organization={IEEE}
}

@article{liu20253d,
  title={3d skeleton-based action recognition: A review},
  author={Liu, Mengyuan and Liu, Hong and Hu, Qianshuo and Ren, Bin and Yuan, Junsong and Lin, Jiaying and Wen, Jiajun},
  journal={arXiv preprint arXiv:2506.00915},
  year={2025}
}

@inproceedings{yin2025recognizing,
  title={Recognizing Skeleton-Based Actions As Points},
  author={Yin, Baiqiao and Lin, Jiaying and Wen, Jiajun and Li, Yue and Liu, Jinfu and Wang, Yanfei and Liu, Mengyuan},
  booktitle={2025 IEEE/RSJ International Conference on Intelligent Robots and Systems (IROS)},
  pages={21582--21589},
  year={2025},
  organization={IEEE}
}

\newpage

\appendix

\renewcommand{\thefigure}{S\arabic{figure}}
\renewcommand{\thetable}{S\arabic{table}}

\setcounter{figure}{0}
\setcounter{table}{0}

\section*{Supplementary Material}
\setcounter{section}{0}
\renewcommand{\thesection}{\Alph{section}}

In this supplementary material, we provide a comprehensive overview of the UniMM-HAR dataset, an analysis of its heterogeneous multi-source characteristics, an investigation of Doppler properties as a motion prior, and additional experiments and visualizations that demonstrate the superiority and interpretability of DAP-Net on heterogeneous data, complementing the main paper.
The supplementary material is organized as follows:
\begin{itemize}
    \item Section~\ref{sec:Dataset}: Details of the UniMM-HAR Dataset
    \item Section~\ref{sec:heterogeneity_analysis}: Heterogeneity Analysis
    \item Section~\ref{sec:doppler}: Doppler Analysis
    \item Section~\ref{sec:Experiment}: Additional Experiments and Visualizations
\end{itemize}

\section{Details of UniMM-HAR Dataset}
\label{sec:Dataset}

\begin{figure}
    \centering
    \includegraphics[width=1\linewidth]{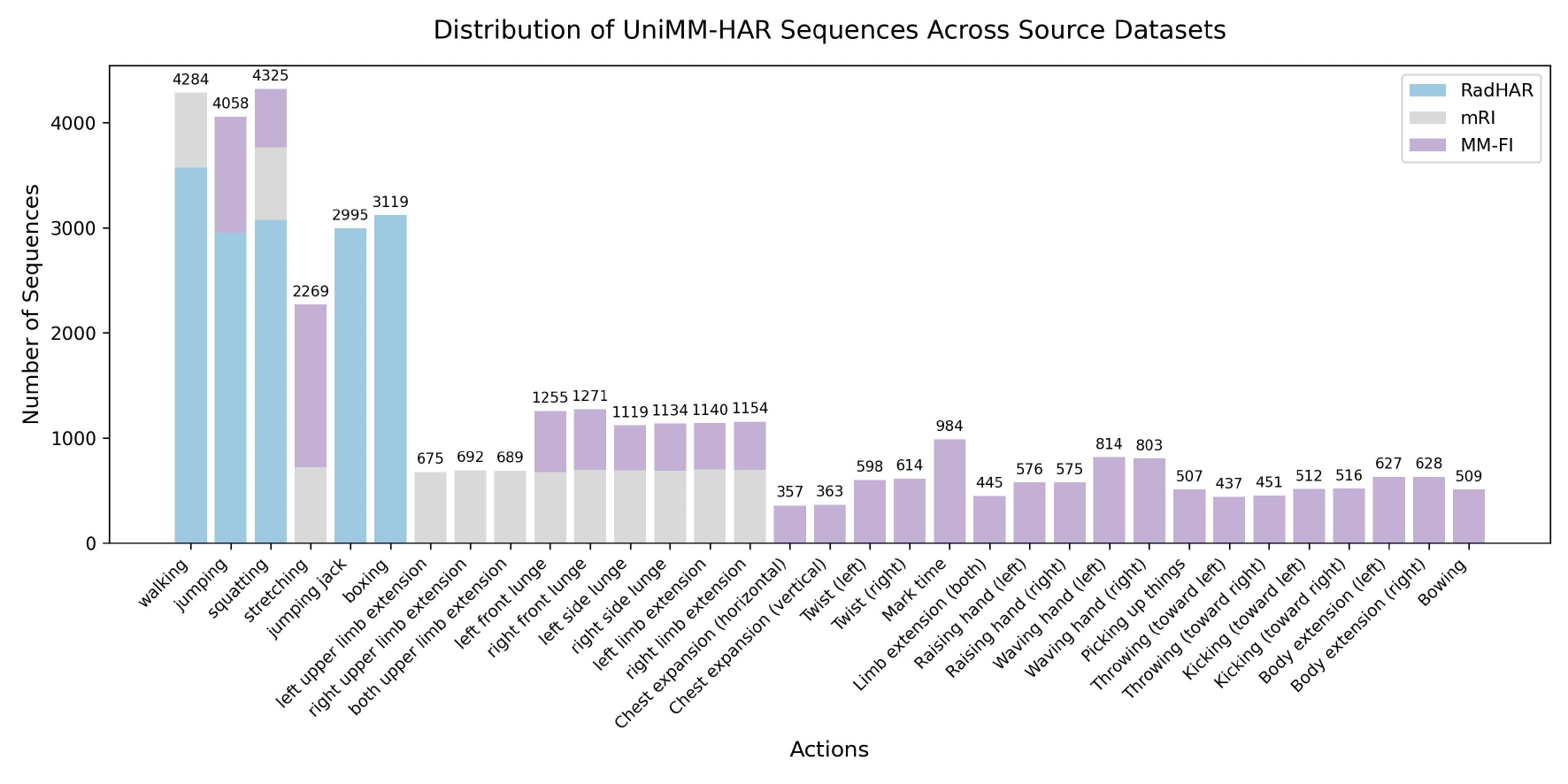}
    \caption{Number of samples per action class, illustrated as a stacked bar chart to show the contributions of different data sources.}
    \label{fig:action_bar}
\end{figure}

The UniMM-HAR dataset provides millimeter-wave (mmWave) point cloud data for human action recognition (HAR), encompassing 33 action categories, including 21 daily activities and 12 rehabilitation exercises. It involves 62 subjects, 40,494 sequences, and over 1.29 million frames. Table~\ref{tab:action_source_counts} summarizes the action categories, the number of sequences per category, and the composition of data sources. 
Fig.~\ref{fig:action_bar} illustrates the distribution of sequences across the different action classes.

\begin{table}[t]
\centering
\caption{Action and cross-source composition of UniMM-HAR. 
For each unified action category, the \textit{Source} column reports the corresponding original action labels in the source datasets along with the number of sequences contributed by each dataset. 
The \textit{Total} column summarizes the aggregated sequence count across all sources. 
Actions are grouped into two semantic categories: \textit{Daily} and \textit{Rehabilitation (Rehab)} actions.}
\label{tab:action_source_counts}
\setlength{\tabcolsep}{8pt}
\renewcommand{\arraystretch}{1.2}
\resizebox{\linewidth}{!}{
\begin{tabular}{r l c c l}
\toprule
\textbf{ID} & \textbf{Action} & \textbf{Source (action / \#seq)} & \textbf{Total} & \textbf{Type} \\
\midrule
0  & walk                    & RadHAR: walking (3574), mRI: walking in a straight line (710)                                   & 4284 & Daily \\
1  & jump                    & RadHAR: jumping (2953), MM-Fi: Jumping up (1105)                                                   & 4058 & Daily \\
2  & squat                   & RadHAR: squatting (3074), mRI: squat (695), MM-Fi: Squat (556)                                      & 4325 & Daily \\
3  & stretch                 & MM-Fi: Stretching and relaxing (1547),  mRI: stretching and relaxing in free forms (722)            & 2269 & Rehab \\
4  & jumping jack            & RadHAR: jumping jack (2995)                                                                         & 2995 & Daily \\
5  & box                     & RadHAR: boxing (3119)                                                                               & 3119 & Daily \\
6  & extend left upper limb  & mRI: left upper limb extension (675)                                                                & 675  & Rehab \\
7  & extend right upper limb & mRI: right upper limb extension (692)                                                               & 692  & Rehab \\
8  & extend both upper limbs & mRI: both upper limb extension (689)                                                                & 689  & Rehab \\
9  & left front lunge        & mRI: left front lunge (675), MM-Fi: Lunge (toward left-front) (580)                                 & 1255 & Rehab \\
10 & right front lunge       & mRI: right front lunge (696), MM-Fi: Lunge (toward right-front) (575)                              & 1271 & Rehab \\
11 & left side lunge         & mRI: left side lunge (693), MM-Fi: Lunge (toward left) (426)                                       & 1119 & Rehab \\
12 & right side lunge        & mRI: right side lunge (688), MM-Fi: Lunge (toward right) (446)                                     & 1134 & Rehab \\
13 & extend left limb        & mRI: left limb extension (701), MM-Fi: Limb extension (left) (439)                                  & 1140 & Daily \\
14 & extend right limb       & mRI: right limb extension (696), MM-Fi: Limb extension (right) (458)                                & 1154 & Daily \\
15 & expand chest horizontally & MM-Fi: Chest expansion (horizontal) (357)                                                          & 357  & Daily \\
16 & expand chest vertically   & MM-Fi: Chest expansion (vertical) (363)                                                            & 363  & Daily \\
17 & twist left             & MM-Fi: Twist (left) (598)                                                                           & 598  & Daily \\
18 & twist right            & MM-Fi: Twist (right) (614)                                                                          & 614  & Daily \\
19 & mark time              & MM-Fi: Mark time (984)                                                                              & 984  & Rehab \\
20 & extend both limbs      & MM-Fi: Limb extension (both) (445)                                                                  & 445  & Rehab \\
21 & raise left hand        & MM-Fi: Raising hand (left) (576)                                                                    & 576  & Daily \\
22 & raise right hand       & MM-Fi: Raising hand (right) (575)                                                                   & 575  & Daily \\
23 & wave left hand         & MM-Fi: Waving hand (left) (814)                                                                     & 814  & Daily \\
24 & wave right hand        & MM-Fi: Waving hand (right) (803)                                                                    & 803  & Daily \\
25 & pick up object         & MM-Fi: Picking up things (507)                                                                      & 507  & Daily \\
26 & throw left             & MM-Fi: Throwing (toward left) (437)                                                                 & 437  & Daily \\
27 & throw right            & MM-Fi: Throwing (toward right) (451)                                                                & 451  & Daily \\
28 & kick left              & MM-Fi: Kicking (toward left) (512)                                                                  & 512  & Daily \\
29 & kick right             & MM-Fi: Kicking (toward right) (516)                                                                 & 516  & Daily \\
30 & extend left body       & MM-Fi: Body extension (left) (627)                                                                  & 627  & Rehab \\
31 & extend right body      & MM-Fi: Body extension (right) (628)                                                                 & 628  & Rehab \\
32 & bow                    & MM-Fi: Bowing (509)                                                                                 & 509  & Daily \\
\bottomrule
\end{tabular}%
}
\end{table}

\subsection{Source Datasets}

The UniMM-HAR dataset is constructed by harmonizing three representative open-source mmWave radar benchmarks: RadHAR \cite{singh2019radhar}, mRI \cite{mRI}, and MM-Fi \cite{mmfi}. These datasets exhibit significant heterogeneity in terms of radar hardware, frequency bands, acquisition parameters, and experimental environments, providing a challenging foundation for evaluating cross-source generalization.
\begin{itemize}
    \item \textbf{RadHAR} \cite{singh2019radhar} is a foundational dataset specifically designed for human action recognition using mmWave radar. The data were collected using a TI IWR1443BOOST radar \cite{ti_iwr1443boost}, operating in the 76--81\,GHz band and employing an FMCW (Frequency-Modulated Continuous-Wave) waveform. The dataset captures five daily action categories (e.g., walking, jumping, and falling) performed by two subjects in an indoor environment.
    \item \textbf{mRI} \cite{mRI} is a comprehensive multi-modal benchmark intended for 3D human pose estimation and action recognition. It utilizes a TI IWR1443 BOOST radar \cite{ti_iwr1443boost} at 77 GHz to capture 12 clinically relevant movements (e.g., squatting, lunge, limb). The data was collected from 20 subjects and included synchronized 3D point clouds alongside RGB-D and inertial sensor data. Compared to other sources, mRI contributes highly diverse limb-level motion patterns, offering a rich variety of kinematic variations essential for fine-grained action modeling.
    \item \textbf{MM-Fi} \cite{mmfi} is a large-scale multi-modal non-intrusive 4D human dataset designed to bridge the gap between wireless sensing and high-level perception tasks. It employs a TI IWR6843 radar \cite{ti_iwr6843} operating in the 60-64 GHz band. The dataset is notably diverse, covering 27 daily or rehabilitation action categories performed by 40 subjects across 4 distinct physical scenes. As the first dataset to provide five non-intrusive modalities for 4D human pose estimation, its radar stream is synchronized with LiDAR and RGB-D sensors, providing a high-fidelity reference for complex activity recognition in varied environments.
\end{itemize}

\subsection{Dataset Processing Details}
\subsubsection{Action Alignment}
Tab.~\ref{tab:action_source_counts} presents the action alignment and merging processes across different datasets. For semantically equivalent actions, we unify them into a single action category. For example, \textit{RadHAR: squatting}, \textit{mRI: squat}, and \textit{MM-Fi: Squat} are unified as \textit{squat}, while \textit{MM-Fi: Stretching and relaxing} and \textit{mRI: stretching and relaxing in free forms} are merged into \textit{stretch}.
To preserve finer-grained action differences, actions with subtle semantic distinctions are retained as separate categories rather than being directly merged. For instance, \textit{MM-Fi: Limb extension (both)} is retained as \textit{extend both limbs}, whereas \textit{mRI: both upper limb extension} is kept as \textit{extend both upper limbs}. Similarly, \textit{extend left upper limb} and \textit{extend left limb} are distinguished as different actions.
This action alignment strategy balances semantic consistency and action granularity across datasets, enabling the model to share knowledge while preserving dataset-specific action variations, thereby facilitating more reliable heterogeneous learning and evaluation under cross-source distributions.

\subsubsection{Released Data Versions}
For flexibility and reproducibility, the dataset is released in two formats: CSV and NPZ. 
The CSV files preserve data that are close to the original distributions, while the NPZ files provide standardized representations that can be directly used for model. 
Each sample follows the naming convention $D_iA_jE_kP_mS_n$, e.g., \textit{D001A001E001P001S0001},
where $D$ denotes the dataset identifier, $A$ the action category, $E$ the acquisition environment, $P$ the participant index, and $S$ the sample index.
This naming scheme explicitly encodes the dataset source, action category, acquisition scene, and subject identity of each sequence.

\subsubsection{CSV Format: Dataset-Aware Preprocessing}
We adopt dataset-specific preprocessing strategies to construct initial point cloud sequence samples for the three datasets, which are subsequently saved in CSV format. Each CSV file contains the following fields: \textit{Frame, X, Y, Z, Doppler, Intensity}, representing the frame index, the 3D coordinates of each point, the Doppler velocity, and the radar reflection intensity, respectively.
\begin{itemize}
    \item \textbf{For the RadHAR dataset}, each file corresponds to a long millimeter-wave point cloud sequence. Following the official preprocessing pipeline, we employ a sliding window approach to extract sub-sequences from the raw sequence, with a window length of 60 frames and a stride of 10 frames. This setting preserves temporal continuity while generating multiple overlapping action segments. The resulting point cloud sequences are stored in CSV format for subsequent data processing and model training.
    \item \textbf{For the mRI dataset}, each action class is annotated with a temporal range based on frame indices. We first segment the raw sequences into action-specific sub-sequences according to the annotated frame ranges. Since some action sequences remain relatively long and exhibit significant variations in duration across actions, we further apply a sliding window strategy with a window size of 32 frames and a stride of 16 frames. Given that the mRI dataset has a frame rate of 10\,Hz, each window covers approximately 3.2 seconds of motion, ensuring that longer actions can be fully captured within a single sample. In addition, the 50\% overlap balances the trade-off between generating sufficient training samples and avoiding excessive redundancy.
    \item \textbf{For the MM-Fi dataset}, the raw sequences are not pre-segmented by action, and a single video sequence typically contains multiple consecutive actions. We therefore first split the raw sequences into action-level segments according to the segmentation files provided by the dataset. Since the resulting action sequences are relatively short, no further processing is applied.
\end{itemize}
Note that the preprocessing procedures modify the dataset statistics. 
The sliding-window strategy increases the number of sequence samples by generating overlapping sub-sequences from long recordings, 
which introduces partial frame redundancy across adjacent samples. 
In contrast, action-level segmentation removes frames outside annotated action intervals, 
thereby eliminating irrelevant background segments. 
Consequently, the total number of frames and sequences after preprocessing differs from the original dataset statistics. 
The final statistics used in our benchmark are reported in Table~2 of the main paper.

\subsubsection{NPZ Format:Representation Standardization}
To enable consistent model input, we standardize each point cloud clip to a fixed tensor shape of $[T, P, C] = [32, 64, 5]$, where the five channels correspond to the 3D coordinates $(x, y, z)$, Doppler velocity, and intensity.
For the temporal dimension $T$, if a clip contains more than 32 frames, we apply uniform temporal downsampling to select 32 representative frames; if fewer than 32 frames are available, zero-padding is used to reach the required length.
For the spatial dimension $P$, 64 points per frame are selected, which is sufficient to represent the point counts of most frames in the dataset as shown in Table~\ref{tab:point_density_statistics}.When a frame has more than 64 points, we employ Farthest Point Sampling (FPS) to downsample to exactly 64 points; when it has fewer, we randomly repeat points to reach the target count.  


\subsection{Evaluation Protocols}
\label{sec:Evaluation}

\begin{table}[t]
\centering
\caption{Overview of the mmWave radar datasets used to build UniMM-HAR and the evaluation protocols they support. 
CS and CE indicate the availability of Cross-Subject (C-Sub) and Cross-Set (C-Set) protocols, respectively. 
``N/A '' denotes that the information is not reported in the original dataset. 
\#Samples represents the number of processed action samples included in UniMM-HAR.}
\label{tab:dataset_Protocols}
\resizebox{0.85\linewidth}{!}{\begin{tabular}{lcccccc}
\toprule
\textbf{Dataset} &  \textbf{Protocols} & \textbf{\#Actions} & \textbf{\#Subjects} & \textbf{\#Scenes} & \textbf{\#Samples} \\
\midrule
RadHAR \cite{singh2019radhar}   & N/A      & 5  & 2  & N/A  & 15,715 \\
mRI \cite{mRI}    & CS      & 12 & 20 & N/A  & 8,332 \\
MM-Fi \cite{mmfi}    & CS, CE  & 27 & 40 & 4  & 16,448 \\
\bottomrule
\end{tabular}}

\end{table}

\begin{table}[t]
\centering
\caption{Summary of evaluation protocols in UniMM-HAR.\#Act. refers to the number of action categories. \#Train, \#Test, and \#Total denote the number of samples in the training set, test set, and the entire dataset, respectively. }
\label{tab:protocol_summary}
\resizebox{\linewidth}{!}{
\begin{tabular}{lcccccc}
\toprule
Protocol & Data  Sources & Split  Criterion & \#Act. & \#Train  & \#Test  & \#Total \\
\midrule
Random Split & RadHAR, mRI, MM-Fi & 60:40 (per source) & 33 & 24,297 & 16,197 & 40,494 \\
C-Sub & mRI, MM-Fi & Subject split (5:5) & 25 & 10,053 & 9,604 & 19,657 \\
C-Set & RadHAR, mRI, MM-Fi & Scene split (4:2) & 27 & 19,603 & 8,437 & 28,041 \\
\bottomrule
\end{tabular}

}

\end{table}

As summarized in Table~\ref{tab:dataset_Protocols}, the source datasets in UniMM-HAR support different evaluation protocols. Accordingly, UniMM-HAR provides three evaluation settings for heterogeneous multi-source scenarios: Random Split, Cross-Subject (C-Sub), and Cross-Set (C-Set), as summarized in Table~\ref{tab:protocol_summary}.

\noindent   \textbf{Random Split.}
In this protocol, samples from each data source are randomly split into training and testing sets with a 60:40 ratio, and the splits from all sources are aggregated to form the final training and testing sets. The training set contains 24,297 samples, and the test set contains 16,197 samples.

\noindent \textbf{Cross-Subject (C-Sub).}
This protocol evaluates the model’s ability to generalize across different subjects. As shown in Table~\ref{tab:dataset_Protocols}, the cross-subject evaluation only involves datasets with subject annotations, namely mRI and MM-Fi, while RadHAR is excluded. The cross-subject subset contains 19,657 samples spanning 25 action categories. The training set consists of mRI subjects 0–9 and MM-Fi subjects 0–19, totaling 10,053 samples. The test set includes mRI subjects 10–19 and MM-Fi subjects 20–39, totaling 9,604 samples.

\noindent \textbf{Cross-Set (C-Set).}
This protocol evaluates generalization across different acquisition scenes. It includes data from RadHAR, mRI, and MM-Fi, which differ in collection scenes. The cross-set subset contains 28,041 sequences covering 27 action categories across six scenes. The data are split into training and testing sets according to a 4:2 scene ratio. The training set includes MM-Fi subjects 2–3, all RadHAR subjects, and all mRI subjects, totaling 19,603 samples. The test set includes MM-Fi subjects 0–1, totaling 8,437 samples.

\subsection{Visualization}

\begin{figure}
    \centering
    \includegraphics[width=1\linewidth]{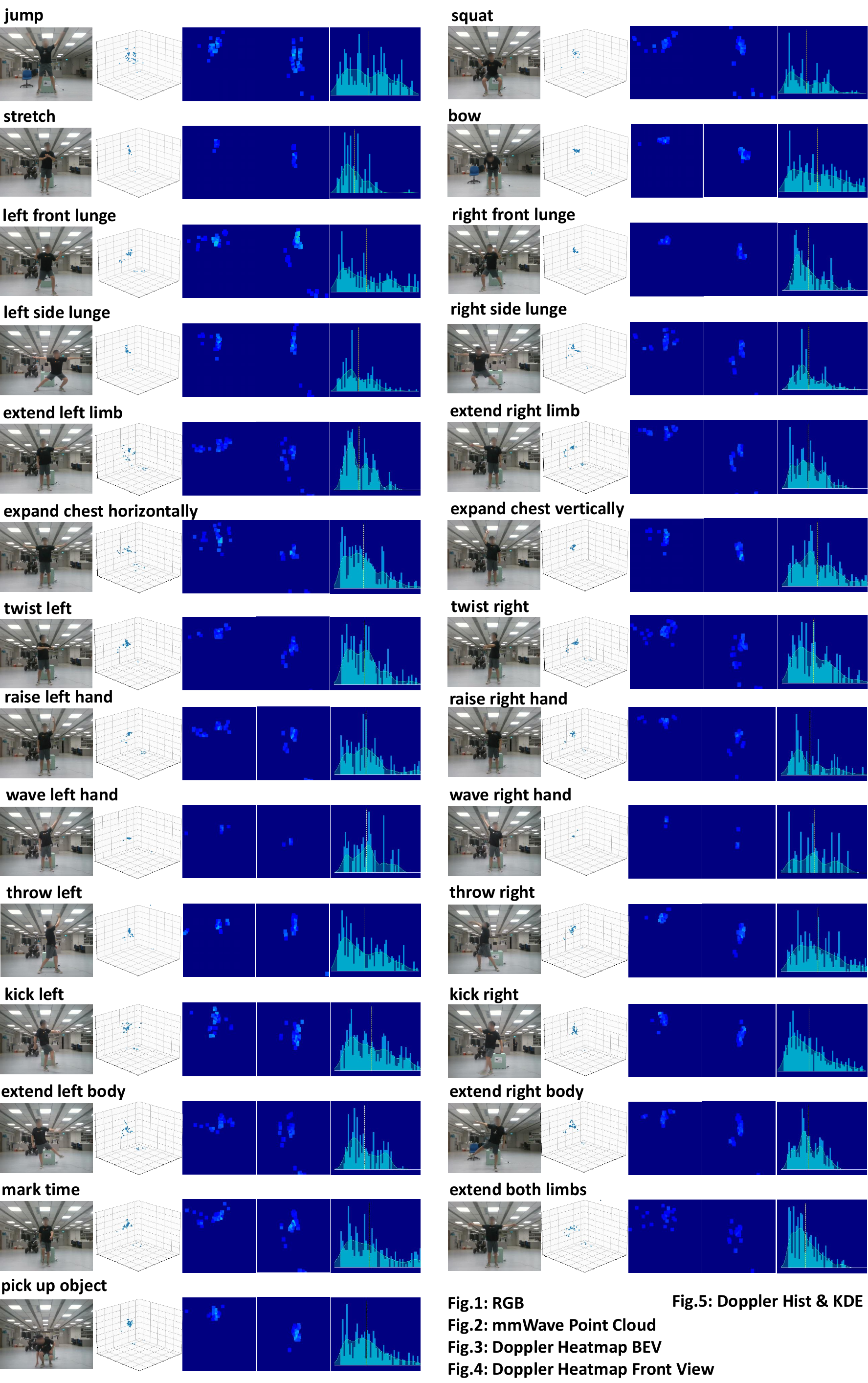}
    \vspace{-5mm}
    \caption{Samples visualization of different action categories in the UniMM-HAR dataset. Each sample includes 5 visualizations.
    }
    \label{fig:dataset_visulization}
\end{figure}

Fig.~\ref{fig:dataset_visulization} illustrates visualizations of representative samples in the dataset. 
Each sample includes five modalities, arranged from left to right: the RGB image, the mmWave point cloud, Doppler heatmaps in bird’s-eye view (BEV) and front view, and the Doppler value distribution. 
In the Doppler heatmaps, brighter colors indicate larger Doppler values, typically corresponding to stronger or faster motion. 
The Doppler value distribution panel shows the histogram and kernel density estimation (KDE) for each action category, with the horizontal axis representing Doppler values and the vertical axis representing normalized probability density.
The blue bars denote the histogram, the green curve and shaded region correspond to the KDE distribution, and the yellow dashed line indicates the mean Doppler value.
From these visualizations, several observations can be made. First, \uline{mmWave point clouds are typically sparse and cannot directly reveal clear human shapes, which provides a degree of privacy preservation but also increases the difficulty of action modeling.} Second, \uline{the Doppler heatmaps show that larger Doppler values tend to concentrate in regions with more pronounced motion, reflecting the velocity patterns of human movements.} Furthermore, the Doppler distributions vary across different action categories, suggesting that \uline{Doppler information provides discriminative motion cues for HAR}.

\section{Heterogeneity Analysis}
\label{sec:heterogeneity_analysis}
This section provides a detailed analysis of the heterogeneity in the UniMM-HAR dataset, covering both the sources of heterogeneity in Section~\ref{sec:Heterogeneity_Sources} and its impact arising from such heterogeneity in Section~\ref{sec:Impact_Heterogeneity}. 
The effectiveness of DAP-Net in heterogeneous cross-source scenarios is presented in Section~\ref{sec:Heterogeneous_Evaluation}.

\subsection{Heterogeneity Sources}
\label{sec:Heterogeneity_Sources}

\begin{table}[t]
\centering
\caption{Heterogeneity of Radar Devices and Recording Setups in UniMM-HAR}
\label{tab:dataset_summary}

\resizebox{\textwidth}{!}{%
\begin{tabular}{l l l l c}
\toprule
\textbf{Dataset} &
\makecell{\textbf{Radar} \\ \textbf{Device}} &
\makecell{\textbf{Frequency} \\ \textbf{Band}} &
\makecell{\textbf{Frame} \\ \textbf{Rate}} &
\textbf{Setup} \\
\midrule

RadHAR & TI IWR1443 & 76--81 GHz & 30 Hz &
Radar mounted on a tripod at 1.3 m height \\

mRI & TI IWR1443 & 76--81 GHz & 10 Hz &
Radar placed on a table 2.4 m from the subject \\

MM-Fi & TI IWR6843 & 60--64 GHz & 30 Hz &
All subjects positioned 3.0 m away from the platform \\

\bottomrule
\end{tabular}%
}
\end{table}

\begin{table}[t]
\centering
\caption{Frame-wise point count statistics across data sources in UniMM-HAR.}
\begin{tabular}{lcccccccc}
\toprule
Source & Mean & Std & Min & 25\% & 50\% & 75\% & Max \\
\midrule
RadHAR &  21.42 & 3.90  & 7  & 19 & 21 & 24 & 42  \\
mRI    &  29.89 & 17.53 & 1  & 16 & 29 & 43 & 64  \\
MM-Fi  & 31.33 & 17.16 & 1  & 18 & 30 & 43 & 112 \\
\bottomrule
\end{tabular}
\label{tab:point_density_statistics}
\end{table}

UniMM-HAR is composed of three data sources (RadHAR, mRI, and MM-Fi), which differ in radar device, operating frequency, frame rate, and acquisition geometry as Table~\ref{tab:dataset_summary}. These differences introduce heterogeneity into the dataset.

\subsubsection{Device and Frequency Differences}
RadHAR and mRI use the TI IWR1443 radar (76–81~GHz), whereas MM-Fi uses the TI IWR6843 radar (60–64~GHz). 
These differences in radar hardware and carrier frequency induce variations in Doppler measurements and angular resolution, which in turn affect the point cloud feature distributions, causing source-dependent differences in both velocity-related and spatial characteristics. 
Specifically, the carrier frequency \(f_c\) determines the Doppler shift \(f_d\) and angular resolution \(\theta\):
\begin{equation}
    f_d = \frac{2 v f_c}{c}, \quad
    \theta \approx \frac{\lambda}{D} = \frac{c}{f_c D},
\end{equation}
where \(v\) is the target velocity, \(c\) is the speed of light, \(\lambda\) is the wavelength, and \(D\) is the antenna aperture. 
For a given target motion, higher carrier frequencies produce larger Doppler shifts and finer angular resolution, yielding higher-resolution velocity measurements and more detailed spatial structures in azimuth and elevation. Conversely, lower-frequency radars result in smaller Doppler shifts and coarser angular resolution, which can introduce greater discretization in both Doppler and spatial domains of the point clouds.

\subsubsection{Temporal Sampling Differences}
The frame rates of the data sources vary: RadHAR and at MM-Fi 30~Hz, mRI  at 10~Hz. The frame rate \(f_s\) determines temporal sampling density. According to the Nyquist sampling theorem, to capture the highest frequency component \(f_{\text{max}}\) of human motion without aliasing, it is required that \(f_s \ge 2 f_{\text{max}}\). Lower frame rates (10~Hz) result in sparser temporal trajectories, potentially failing to resolve fast motions and thereby affecting the temporal continuity and completeness of motion patterns in the point clouds.

\subsubsection{Acquisition Geometry Differences}
\begin{figure}
    \centering
    \includegraphics[width=1\linewidth]{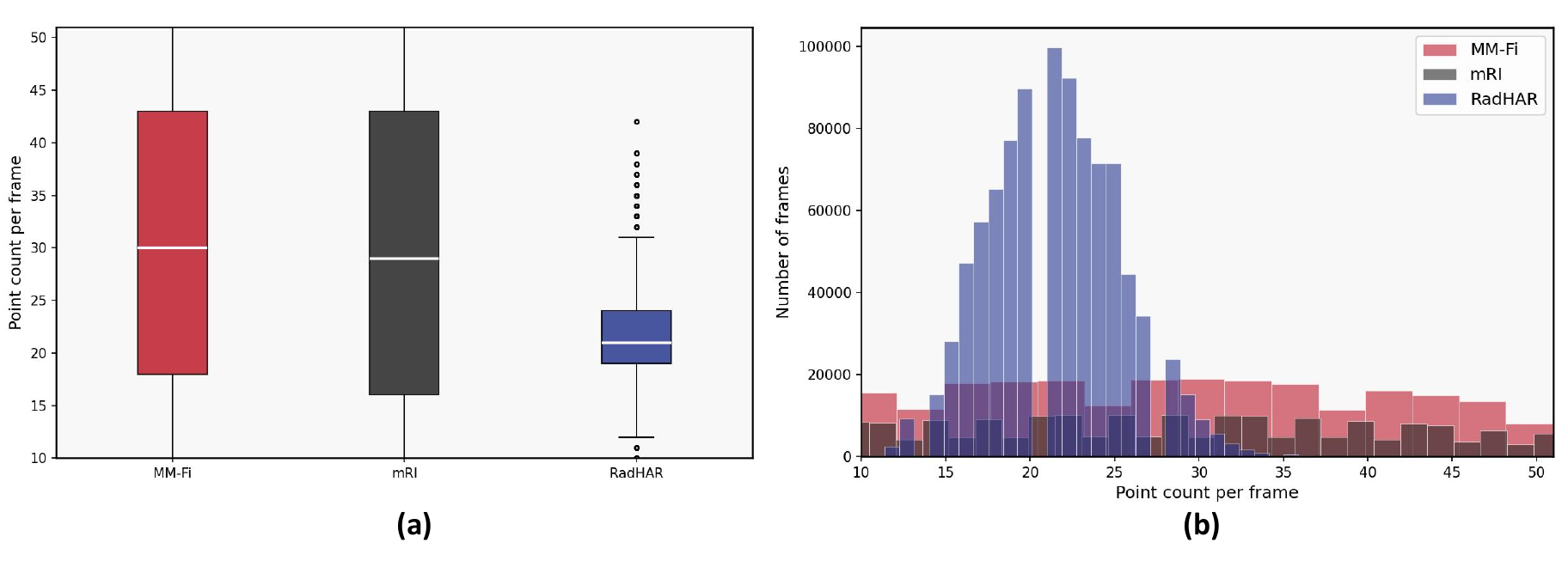}
    \caption{Distribution of frame-wise point counts across heterogeneous data sources.
(a) Box plot of point count distributions.
(b) Histogram of frame-wise point counts aggregated over all frames.
To reduce the influence of extreme outliers, the axis range is restricted to the 5th–95th percentile of the distribution.}
    \label{fig:Heterogeneity_point}
\end{figure}

The radar mounting heights and subject distances differ across datasets. According to the radar equation:
\begin{equation}
    P_r \propto \frac{P_t G_t G_r \lambda^2 \sigma}{(4\pi)^3 R^4},
\end{equation}
where \(P_r\) is the received power and \(R\) is the target distance, the received signal strength decays with the fourth power of distance; longer distances reduce point detection probability and thus decrease point cloud density. Additionally, mounting height affects the radar's elevation coverage, which changes the observable human body parts (e.g., torso vs. limbs), leading to variations in the spatial coverage and point density distribution of the point clouds.

Collectively, \uline{heterogeneity introduces non-additive, source-specific variations in point cloud density, spatial coverage, and feature distributions (x, y, z, Doppler, intensity).} Therefore, effective cross-source generalization requires explicit consideration of data heterogeneity in both model design and evaluation.

\subsection{Impact of Heterogeneity}
\label{sec:Impact_Heterogeneity}

\subsubsection{Heterogeneity in Point Cloud Density}

\begin{figure}[t]
    \centering
    \includegraphics[width=1\linewidth]{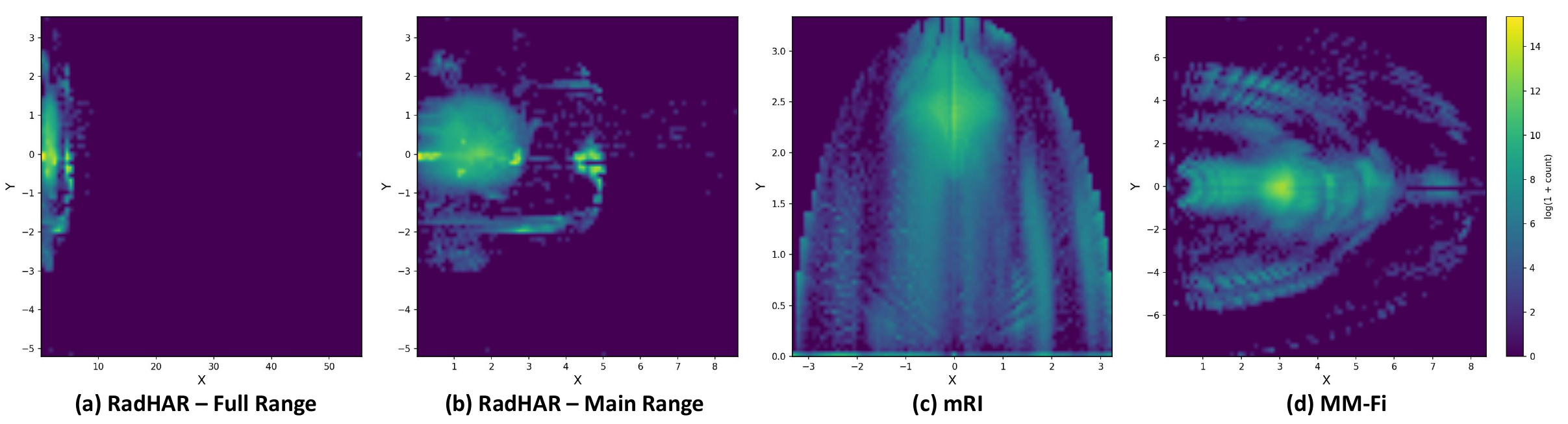}
    \caption{Heatmap visualization of accumulated point cloud projections on the X–Y plane for different data sources. Due to the presence of noticeable outliers in RadHAR, two visualizations are provided: (a) RadHAR – Full Range: includes all points, illustrating the presence of sparse outliers; (b) RadHAR – Main Range: highlights the primary spatial distribution after removing extreme outliers.}
    \label{fig:Heterogeneity_cover}
\end{figure}

To analyze the impact of heterogeneous multi-source data on point cloud density, we compute the frame-wise point count distributions for each source in UniMM-HAR. The visualization within the 5\%–95\% percentile range is shown in Fig.~\ref{fig:Heterogeneity_point}, and the detailed statistics are summarized in Table~\ref{tab:point_density_statistics}.
First, \textbf{the average point density differs noticeably across datasets}. As shown in Table~\ref{tab:point_density_statistics}, MM-Fi and mRI exhibit similar average numbers of points per frame, while RadHAR has a significantly lower mean value of 21.42, indicating that the point clouds in RadHAR are overall sparser than those in the other two datasets.
Second, \textbf{the variability of point density also differs across sources}. MM-Fi and mRI present relatively large standard deviations, suggesting that the number of detected points fluctuates substantially across frames. In contrast, RadHAR shows a much smaller standard deviation, indicating a more stable point density distribution. This difference is also evident in the box plots in Fig.~\ref{fig:Heterogeneity_point}(a), where the distributions of MM-Fi and mRI are considerably wider, while that of RadHAR is more concentrated.
In addition, \textbf{the dynamic range of point density varies across datasets}. MM-Fi and mRI span wide ranges of point counts (1–112 and 1–64, respectively), covering both extremely sparse frames and relatively dense frames. In contrast, the point counts in RadHAR mainly fall within a narrower interval of 7–42, resulting in a more compact distribution. The histogram in Fig.~\ref{fig:Heterogeneity_point}(b) further illustrates these differences.
Overall, these results demonstrate that the datasets exhibit clear heterogeneity in terms of average density, variance, and dynamic range of point clouds. Such differences imply inconsistent spatial sampling characteristics across data sources, which may introduce distribution shifts and pose challenges for unified modeling and cross-source generalization.

\subsubsection{Heterogeneity in Spatial Coverage}

\begin{table}[t]
\centering
\caption{Spatial range statistics of point clouds across different UniMM-HAR data sources. $x_{\text{range}}$, $y_{\text{range}}$, and $z$ indicate the full spatial extent; $x_{95\%}$, $y_{95\%}$, and $z_{95\%}$ show the 95\% coverage of points.}
\begin{tabular}{lcccccc}
\toprule
Source & $x_{\text{range}}$ & $x_{95\%}$ & $y_{\text{range}}$ & $y_{95\%}$ & $z$ & $z_{95\%}$ \\
\midrule
RadHAR & 0.00--55.66 & 0.00--4.84 & -5.20--3.54 & -5.20--0.24 & -8.18--40.51 & -8.18--1.27 \\
mRI    & 3.34--3.23  & -3.34--0.87 & 0.00--3.34 & 0.00--2.81 & -3.34--3.34 & -3.34--1.47 \\
MM-Fi  & 0.03--8.40  & 0.03--3.85 & -7.91--7.91 & -7.91--0.62 & -7.27--7.39 & -7.27--1.07 \\
\bottomrule
\end{tabular}
\label{tab:spatial_range_statistics}
\end{table}

To analyze the spatial distribution differences of point clouds from different sources in the UniMM-HAR dataset, we computed the 3D coordinate ranges and the 95\% point coverage intervals for each source as shown in Table~\ref{tab:spatial_range_statistics}. It can be observed that the RadHAR point clouds exhibit significantly larger extrema in the X, Y, and Z directions compared to mRI and MM-Fi, and the 95\% coverage intervals differ noticeably from the full range, indicating the presence of a small number of outliers. In contrast, the mRI and MM-Fi point clouds are more spatially concentrated, with extrema ranges close to the 95\% coverage intervals, reflecting lower spatial sparsity.
To visually highlight these differences, heatmaps of accumulated point cloud projections on the X–Y plane are presented in Fig.~\ref{fig:Heterogeneity_cover}. Due to the presence of pronounced outliers in RadHAR, two visualizations are provided: Full Range, which includes all points to illustrate sparse outliers, and Main Range, which removes extreme outliers to emphasize the primary spatial distribution. Overall, the differences in spatial coverage, sparsity, and outlier prevalence across data sources are strongly associated with their acquisition geometry and radar installation conditions. Such spatial heterogeneity poses challenges for cross-source modeling and generalization.

\begin{figure}[t]
    \centering
    \includegraphics[width=1\linewidth]{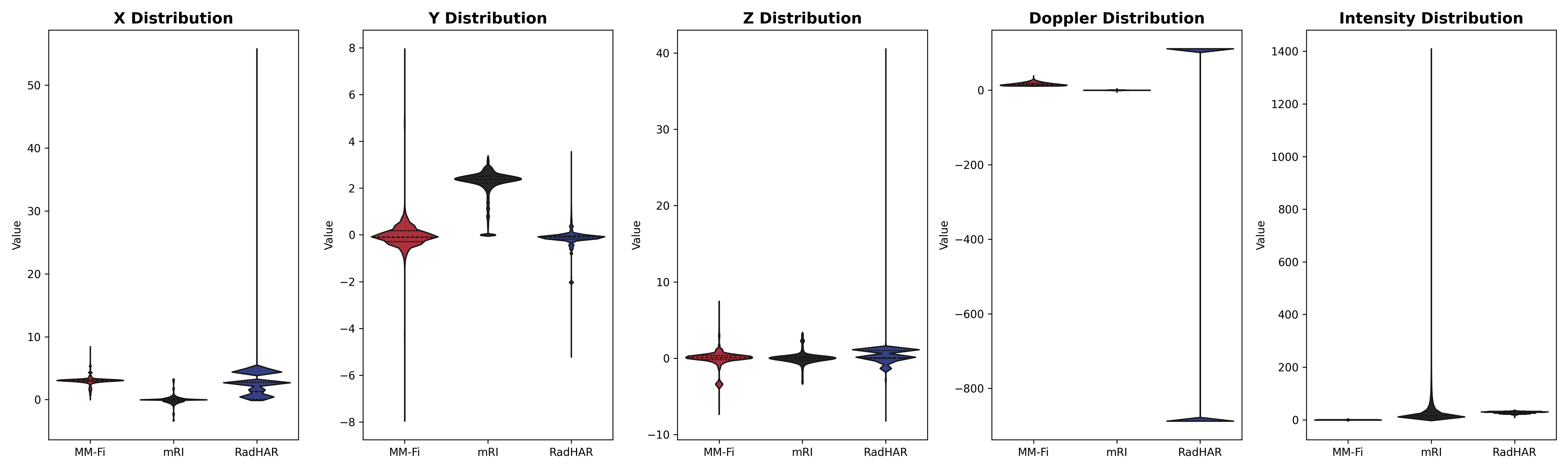}
    \caption{Feature distribution comparison across heterogeneous data sources (X, Y, Z, Doppler, and Intensity).}
    \label{fig:xyzvi}
\end{figure}
\subsubsection{Heterogeneity in Feature Distribution}
We analyzed the feature distributions across three heterogeneous data sources, as shown in Fig.~\ref{fig:xyzvi}. Clear differences can be observed:
\textbf{Spatial Coordinates (X, Y, Z):} The mean values and ranges of spatial coordinates vary systematically across data sources, primarily due to differences in radar devices and acquisition setups.
\textbf{Doppler Velocity:} Doppler distributions exhibit substantial differences across sources. MMFI Doppler values are mostly positive, mRI values are concentrated near zero, and RadHAR contains extreme outliers, while the majority of points still have Doppler values centered around zero.
\textbf{Intensity:} Intensity distributions also differ significantly. MMFI and RadHAR exhibit more concentrated intensity values, whereas mRI shows larger variation.

\section{Doppler Analysis}
\label{sec:doppler}

\uline{Doppler measurements in mmWave radar reflect the radial velocity of a target along the radar line-of-sight and thus provide a direct observation of motion dynamics.} 
For a single scattering point, the Doppler frequency shift is related to the radial velocity as:
\begin{equation}
f_d = \frac{2 v_r f_c}{c},
\end{equation}
where $f_d$ denotes the Doppler frequency shift, $v_r$ is the radial velocity of the target relative to the radar, $f_c$ is the carrier frequency, and $c$ is the speed of light. Accordingly, the radial velocity can be obtained as:
\begin{equation}
v_r = \frac{c\, f_d}{2 f_c}.
\end{equation}
It is important to note that mmWave radar measures only the \textbf{radial velocity} along the line-of-sight direction. Therefore, the observed Doppler reflects only the projection of the true motion velocity onto the radar viewing direction rather than the complete three-dimensional velocity of the target.

\begin{figure}[t]
    \centering
    \includegraphics[width=0.9\linewidth]{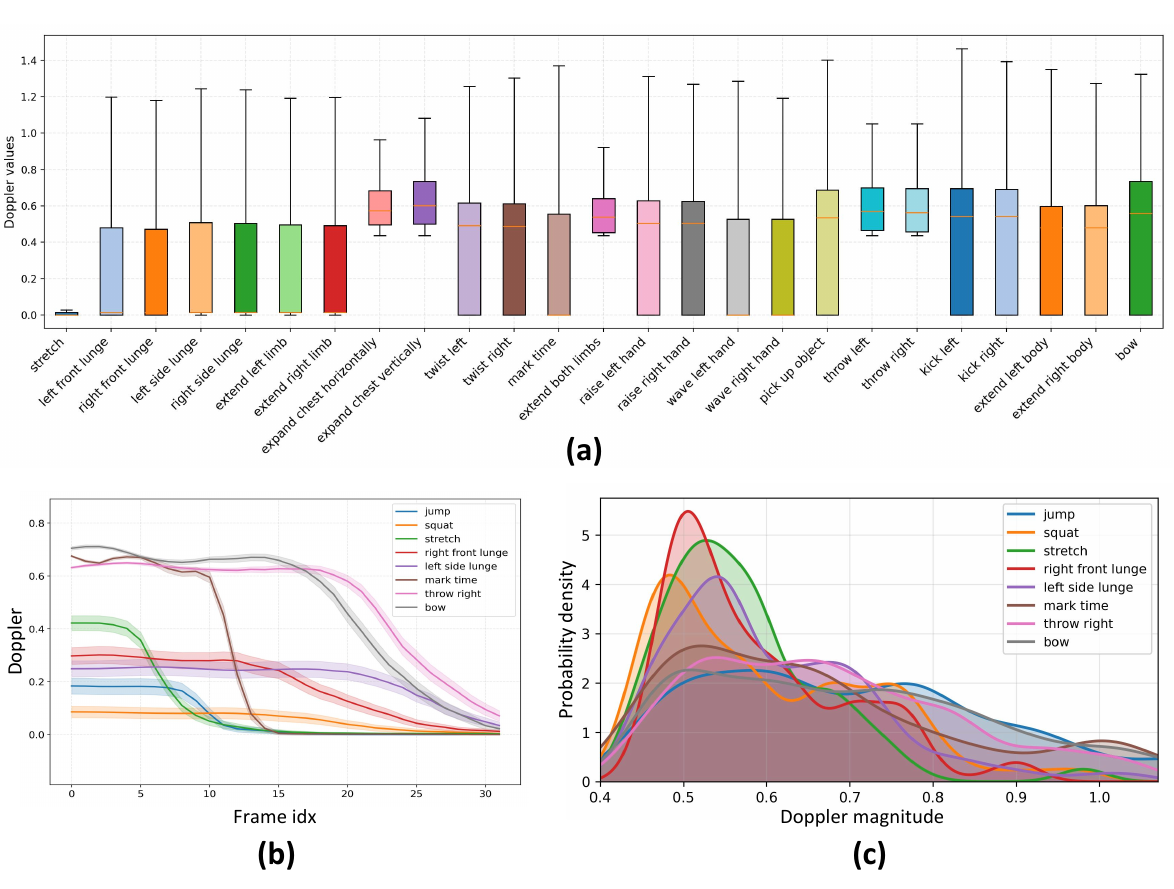}
    \caption{Visualization of Doppler characteristics for different action categories in UniMM-HAR. 
    (a) Box plot showing the distribution of Doppler values. 
    (b) Temporal evolution of Doppler over time. 
    (c) Kernel density estimation (KDE) of Doppler magnitudes.
    }
    \label{fig:doppler}
\end{figure}

Given this physical interpretation, the Doppler values provide meaningful signals for distinguishing different human actions, as variations in motion amplitude and direction manifest in characteristic Doppler patterns. Comparing the RGB images with the Doppler heatmaps in Fig.~\ref{fig:dataset_visulization}, regions with larger motion amplitudes generally exhibit higher Doppler values, serving as a strong attention prior for motion reasoning. Fig.~\ref{fig:doppler} illustrate the Doppler patterns of different actions. It can be observed that the distributions vary significantly across actions and that these variations are semantically meaningful. For instance, the \textit{stretch} action exhibits a small motion amplitude, resulting in relatively low Doppler values, whereas actions with larger movements, such as \textit{expand chest} and \textit{throw}, correspond to higher Doppler readings. Furthermore, actions with similar motion patterns display similar Doppler distributions. For example, \textit{twist left} and \textit{twist right}, which differ only in direction, exhibit comparable distributions. Note that Doppler captures radial velocity along the radar line of sight, so left-right directional differences do not produce significant changes in the Doppler signal.

\uline{In summary, Doppler velocity provides dynamic motion cues that complement spatial geometric information and plays an important role in HAR.} 
In terms of action discriminability, different actions correspond to distinct motion patterns of various human body parts, which are reflected as temporal variations in Doppler velocities. 
These variations encode key motion characteristics such as velocity profiles, periodicity, and acceleration, thereby providing important cues for distinguishing fine-grained actions. 
In terms of cross-source consistency, although heterogeneous radar devices may produce Doppler measurements with different absolute scales, the relative velocity variation patterns generated by the same action, such as the overall evolution trend of Doppler values, tend to remain consistent across devices. 
This stability originates from the intrinsic physical dynamics of human motion and is therefore largely independent of specific sensor parameters.
Therefore, Doppler velocity serves as an effective cue for both action discrimination and cross-source consistency. 
\uline{DAP-Net leverages this Doppler prior to guide the model to focus on motion patterns that are closely related to action semantics while suppressing source-specific interference.}




\section{Additional Experiment and Visualization}
\label{sec:Experiment}

This section primarily includes the hyperparameter \textbf{ablation} in Section~\ref{sec:Ablation}, the \textbf{efficiency} evaluation in Section~\ref{sec:Efficiency}, the \textbf{heterogeneous multi-source evaluation} in Section~\ref{sec:Heterogeneous_Evaluation}, and additional \textbf{visualizations}.

\subsection{Hyperparameter Ablation}
\label{sec:Ablation}

\begin{table}[t]
\centering

\begin{minipage}[t]{0.48\textwidth}
\centering
\caption{Impact of different values of $r$ in $r$-fold duplication.}
\label{tab:rfold}
\resizebox{\linewidth}{!}{
\begin{tabular}{l c c}
\toprule
\textbf{Method} & \textbf{$r$-fold} & \textbf{Acc (\%)} \\
\midrule
DAP-Net (w/o TAM) & 1  & 78.76 \\
DAP-Net (w/o TAM) & 5  & 79.99 $^{\uparrow \textcolor{blue}{1.23}}$ \\
DAP-Net (w/o TAM) & 10 & 79.19 $^{\uparrow \textcolor{blue}{0.43}}$\\
\bottomrule
\end{tabular}
}
\end{minipage}
\hfill
\begin{minipage}[t]{0.48\textwidth}
\centering
\caption{Impact of different target point numbers $P_{\text{goal}}$ after densification.}
\label{tab:pgoal}
\resizebox{\linewidth}{!}{
\begin{tabular}{l c c}
\toprule
\textbf{Method} & \textbf{$P_{\text{goal}}$} & \textbf{Acc (\%)} \\
\midrule
DAP-Net (w/o TAM) & 512  & 79.61 \\
DAP-Net (w/o TAM) & 1024 & 79.99 \\
DAP-Net (w/o TAM) & 2048 & 77.55 \\
\bottomrule
\end{tabular}
}
\end{minipage}

\end{table}

\subsubsection{Effect of $r$-fold Duplication}
In the proposed Tri-branch Motion-Aware Point Densification (TMPD) module, three branches are designed to selectively enhance point cloud representations. The Fast branch employs an $r$-fold duplication strategy to emphasize motion-sensitive points.
We study the impact of the duplication factor $r$ by varying its value. As shown in Table~\ref{tab:rfold}, the performance increases with larger $r$ and saturates around $r=5$, suggesting that \uline{moderate duplication effectively strengthens fast-motion representations while avoiding excessive redundancy.}



\subsubsection{Effect of Target Point Number $P_{\text{goal}}$}
We further analyze the impact of the target point number $P_{\text{goal}}$ in the proposed Doppler-guided Geometry Reparameterization (DGR) module. As described in the main text, DGR performs selective densification guided by per-point Doppler cues and resamples the point cloud to a fixed target size $P_{\text{goal}}$. This process emphasizes motion-discriminative regions while suppressing noisy points, enabling heterogeneous point clouds to form more aligned spatial structures.
To study the effect of the densified point number, we vary $P_{\text{goal}}$ and report the results in Table~\ref{tab:pgoal}. The best performance is achieved when $P_{\text{goal}} = 1024$. When the target size is too small (e.g., 512), the densified point cloud may not sufficiently preserve motion-related structures. In contrast, excessively large values (e.g., 2048) introduce redundant or noisy points, which can degrade recognition performance. \uline{These results indicate that a moderate target point size provides a better balance between structural preservation and noise suppression. Moreover, they demonstrate that the proposed densification strategy is effective and robust when an appropriate target point size is used.}

\subsection{Efficiency Evaluation}
\label{sec:Efficiency}

\begin{table}[t]
\centering

\begin{minipage}[t]{0.45\textwidth} 
\centering
\caption{Overall model complexity comparison with the proposed modules (D²R and TAM).}
\label{tab:complexity_overall}
\resizebox{\linewidth}{!}{%
\begin{tabular}{l c c}
\toprule
\textbf{Backbone} & \textbf{Params (M)} & \textbf{FLOPs (G) }\\
\midrule

PointMLP & 13.24 & 15.75 \\
+D$^2$R & 13.30$^{\color{blue}\uparrow0.06}$ & 15.75$^{\color{blue}\uparrow0.00}$ \\
+D$^2$R+TAM & 14.43$^{\color{blue}\uparrow1.19}$ & 15.84$^{\color{blue}\uparrow0.09}$ \\

\midrule

UST-SSM & 9.19 & 2.79 \\
+D$^2$R  & 9.27$^{\color{blue}\uparrow0.08}$ & 2.79$^{\color{blue}\uparrow0.00}$ \\
+D$^2$R+TAM & 10.03$^{\color{blue}\uparrow0.84}$ & 2.79$^{\color{blue}\uparrow0.00}$ \\

\bottomrule
\end{tabular}%
}
\end{minipage}
\hfill
\begin{minipage}[t]{0.51\textwidth} 
\centering
\caption{Incremental computational cost of the D$^2$R modules (DGR and MFR) under different backbones.}
\label{tab:complexity_increment}
\renewcommand{\arraystretch}{1.4}
\setlength{\tabcolsep}{2pt}
\resizebox{\linewidth}{!}{%
\begin{tabular}{l c c c c}
\toprule
\textbf{Module} & \textbf{Params (M)} & \textbf{Params (\%)} & \textbf{FLOPs (M)} & \textbf{FLOPs (\%)} \\
\midrule

\rowcolor{gray!15}
\multicolumn{5}{l}{\textbf{PointMLP Backbone}} \\

DGR & 0 & 0.000\% & 0 & 0.000\% \\
MFR & 0.06 & 0.505\% & 2.16 & 0.014\% \\

\midrule

\rowcolor{gray!15}
\multicolumn{5}{l}{\textbf{UST-SSM Backbone}} \\

DGR & 0 & 0.000\% & 0 & 0.000\% \\
MFR & 0.06 & 0.650\% & 0.05 & 0.002\% \\

\bottomrule
\end{tabular}%
}
\end{minipage}

\end{table}

Tables \ref{tab:complexity_overall} and \ref{tab:complexity_increment} present the computational complexity analysis of the proposed modules. Table \ref{tab:complexity_overall} reports the overall model complexity under different backbones (PointMLP \cite{ma2022rethinking} and UST-SSM \cite{ustssm}) after incorporating D$^2$R and TAM. As shown, introducing D$^2$R leads to only a marginal increase in parameters while the FLOPs remain nearly unchanged. When TAM is further integrated, the overall computational overhead remains relatively small.
Table \ref{tab:complexity_increment} further analyzes the incremental cost of the two components within D$^2$R, namely DGR and MFR. The results indicate that DGR introduces no additional parameters or FLOPs, while MFR adds only negligible overhead (less than 1\%) in both parameters and computations. \uline{These results demonstrate that the proposed modules enhance the model representation capability while maintaining high computational efficiency.}

\subsection{Heterogeneous Multi-Source Evaluation}
\label{sec:Heterogeneous_Evaluation}

\begin{figure}[t]
    \centering
    \begin{subfigure}[b]{0.49\linewidth}
        \centering
        \includegraphics[width=\linewidth]{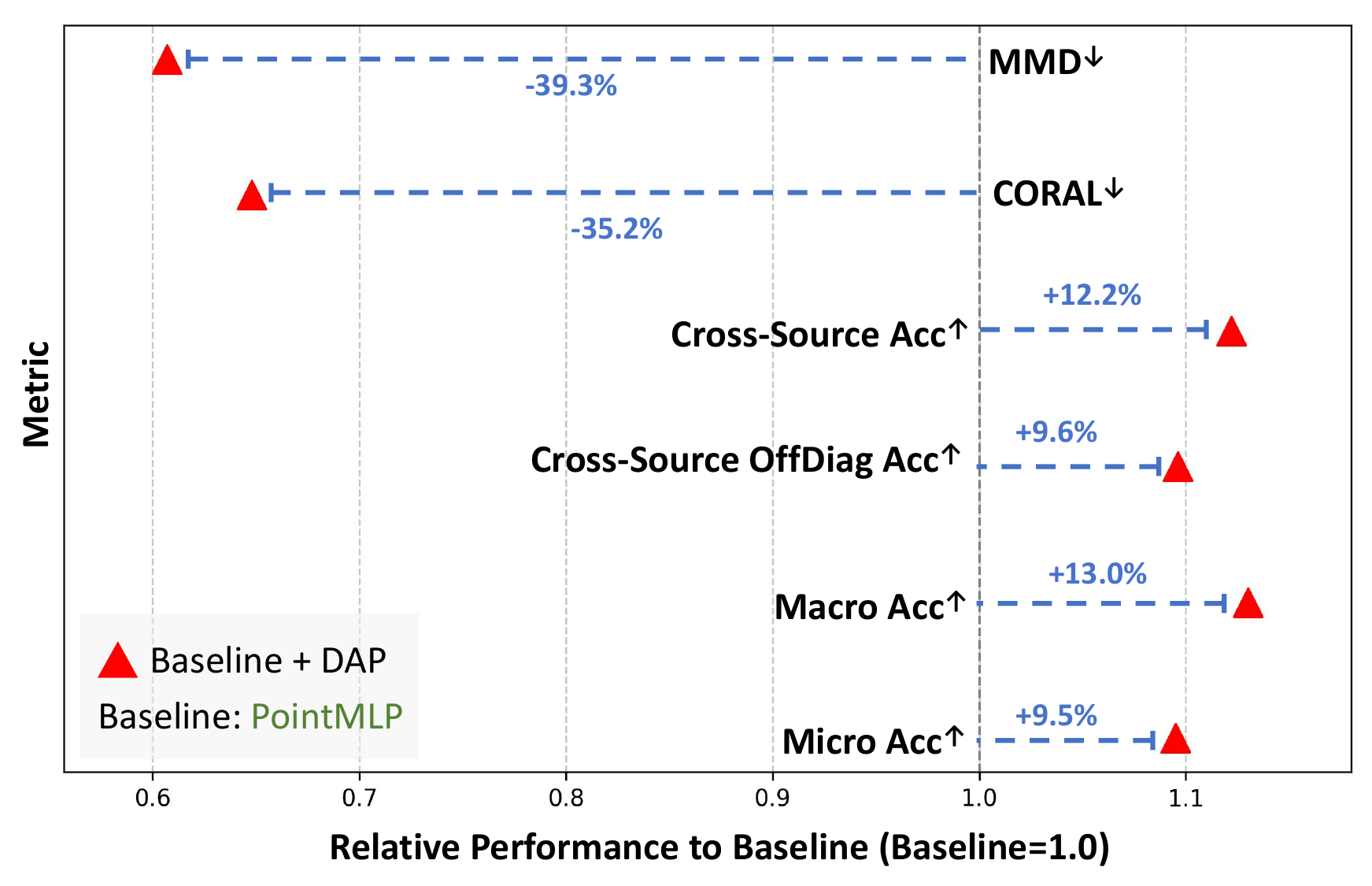}
        \caption{PointMLP vs PointMLP w/ DAP}
        \label{fig:cross_source_pointmlp}
    \end{subfigure}
    \hfill
    \begin{subfigure}[b]{0.49\linewidth}
        \centering
        \includegraphics[width=\linewidth]{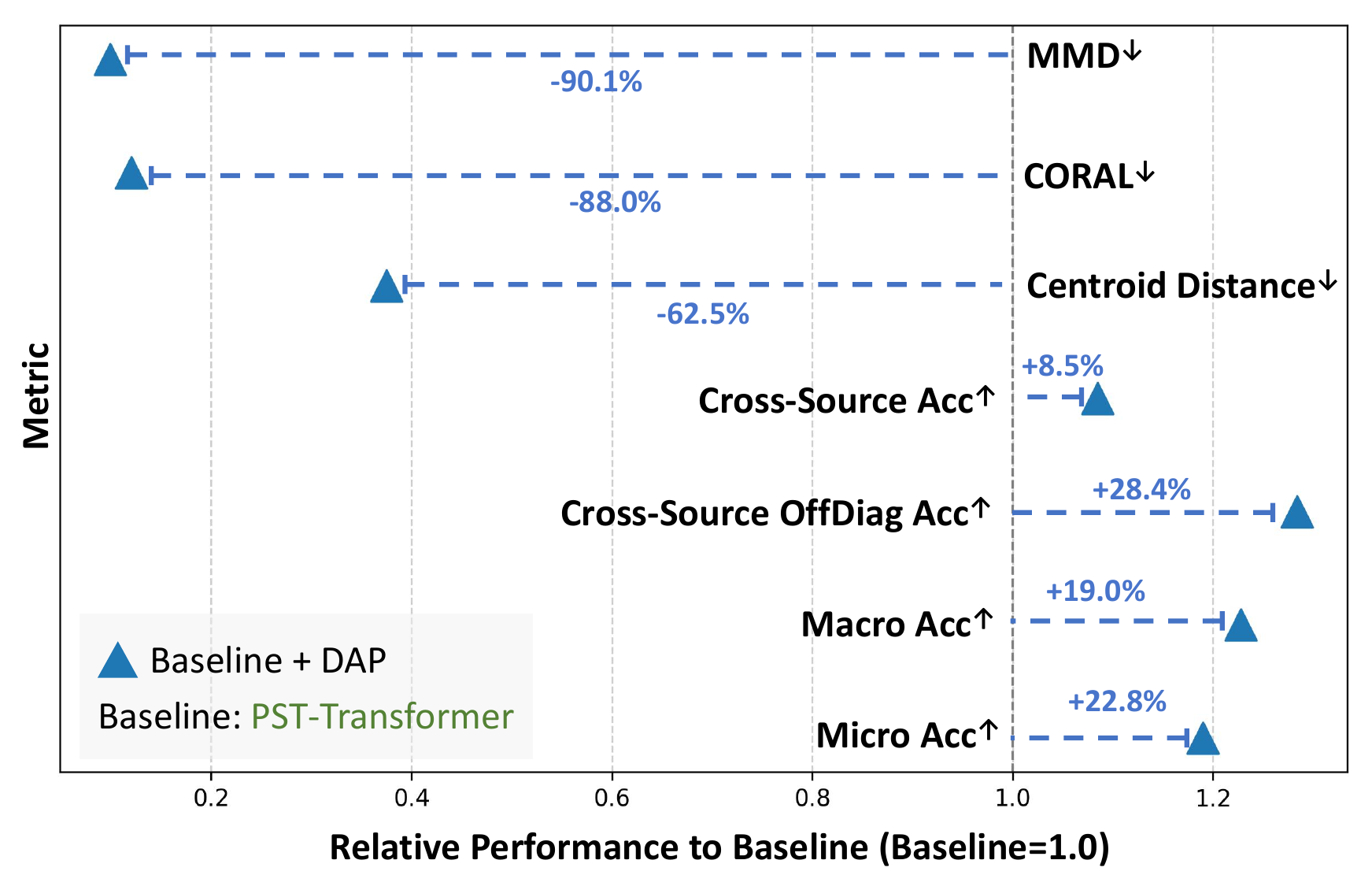}
        \caption{PST-Transformer vs PST w/ DAP}
        \label{fig:cross_source_pst}
    \end{subfigure}
    \caption{Comparison of cross-source performance visualizations. Left: PointMLP baseline improvement; Right: PST-Transformer baseline improvement.}
    \label{fig:cross_source_comparison}
\end{figure}

As analyzed in Section~\ref{sec:heterogeneity_analysis}, UniMM-HAR exhibits significant heterogeneity across sources. To address these distribution shifts, our DAP-Net leverages Doppler as a motion prior, which, as shown in Section~\ref{sec:doppler}, provides a reliable cue for both action discrimination and cross-source consistency. 
DAP-Net enhances modeling in both geometric and feature spaces, using the Doppler prior to guide the network to focus on motion patterns closely related to action semantics while suppressing source-specific interference, and further leverages textual modality to strengthen cross-source representation learning.

To validate the effectiveness of DAP-Net in cross-source heterogeneous scenarios, we conduct experiments on the UniMM-HAR dataset using PointMLP \cite{ma2022rethinking} and PST-Transformer \cite{pst_transformer} as backbones as shown in Fig.~\ref{fig:cross_source_comparison}. The evaluation is performed from three complementary perspectives: \textbf{accuracy metrics} (Macro Accuracy and Micro Accuracy), \textbf{cross-source metrics} (Cross-Source Off-Diagonal Accuracy and Cross-Source Accuracy), and \textbf{source distribution alignment metrics} (Centroid Distance, CORAL, and MMD).

\noindent \textbf{Accuracy metrics} assess the model's performance on both individual classes and overall samples. Specifically, Macro Accuracy quantifies the average prediction accuracy across classes, reflecting the model's capability to maintain balanced performance and mitigating the influence of minority categories, while Micro Accuracy measures the overall correctness across all samples and is less sensitive to class imbalance. Formally, they are defined as:
\begin{equation}
    \text{Macro Acc} = \frac{1}{C}\sum_{c=1}^{C} \frac{\text{TP}_c}{\text{TP}_c + \text{FN}_c}, \quad
    \text{Micro Acc} = \frac{\sum_{c=1}^{C} \text{TP}_c}{\sum_{c=1}^{C} (\text{TP}_c + \text{FN}_c)}.
\end{equation}

\begin{figure}
    \centering
    \includegraphics[width=0.9\linewidth]{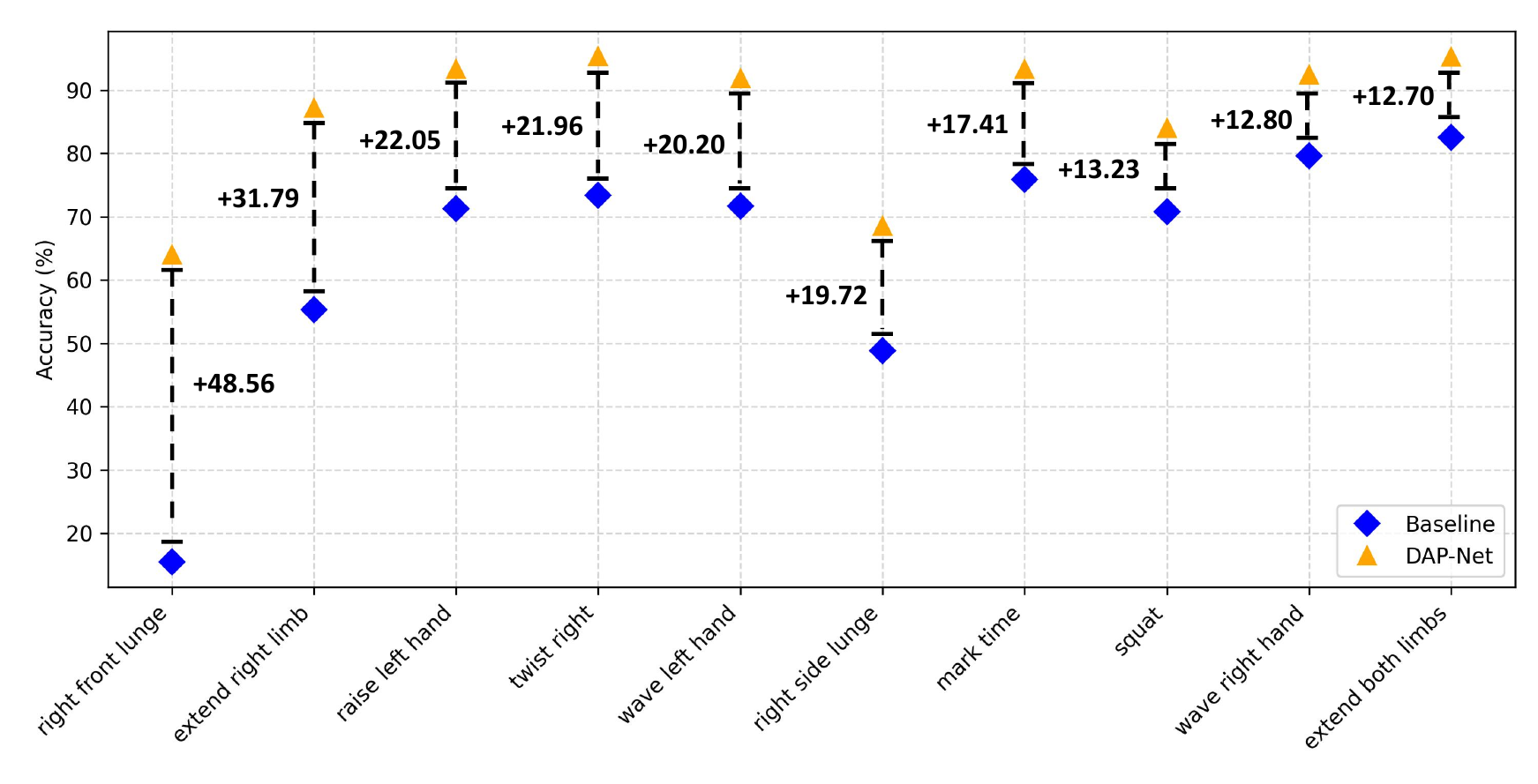}
\caption{Top 10 action categories in UniMM-HAR C-Sub where DAP-Net outperforms the baseline (PointMLP).}
    \label{fig:per_class_acc}
\end{figure}

\noindent  \textbf{Cross-source metrics} evaluate the model's ability to generalize across heterogeneous radar sources. Cross-Source Off-Diagonal Accuracy captures inter-source confusion:
\begin{equation}
    \text{OffDiag Acc} = \frac{\sum_{i \neq j} \mathbf{1}(y_i = \hat{y}_j)}{N_\text{offdiag}},
\end{equation}
quantifying the model's capacity to distinguish action categories between different sources. To further assess cross-source generalization under more challenging conditions, we introduce a strict \textbf{cross-source evaluation protocol} in which nine action classes (A001–A003 and A009–A014) are selected, MM-Fi samples are used as the test set, and only RadHAR and mRI samples are used for training. This setup imposes a more demanding evaluation, as the model must generalize to a completely unseen cross-source distribution, thereby reflecting its robustness to heterogeneous data shifts.

\begin{figure}
    \centering
    \includegraphics[width=1\linewidth]{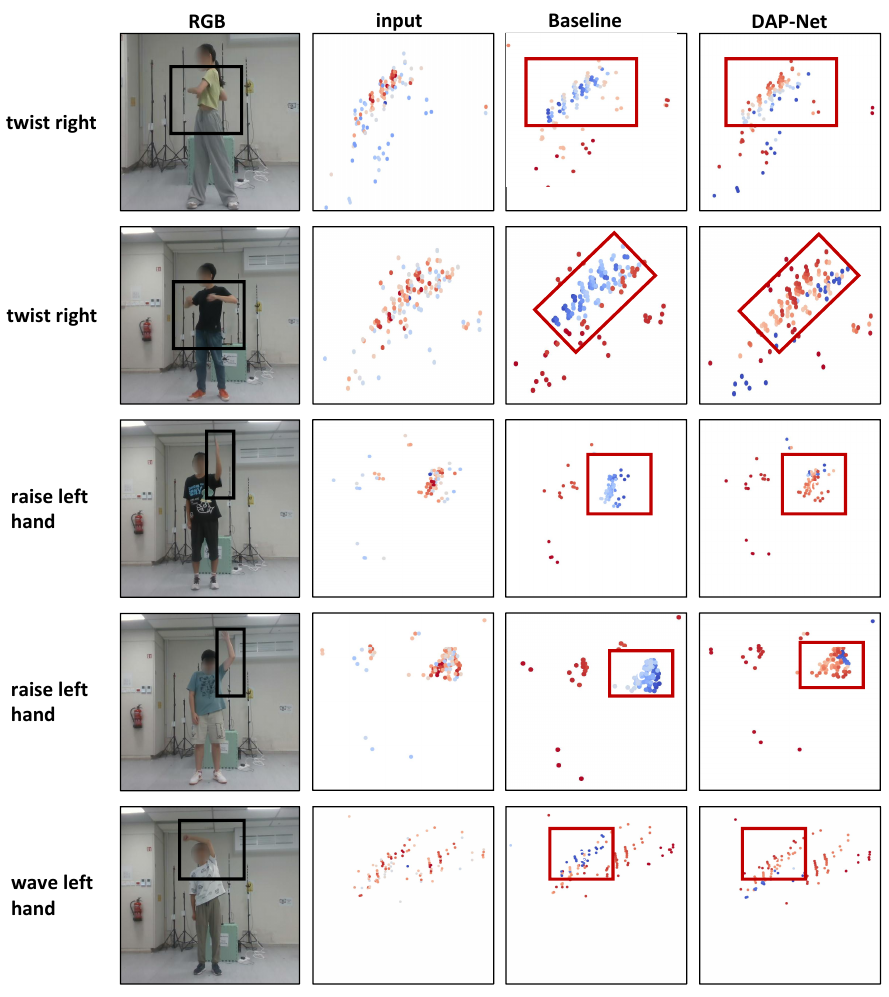}
    \caption{Visualization of point cloud attention. For each action, we show (from left to right): RGB image, input point cloud colored by Doppler velocity, attention weights from baseline model, and attention weights from DAP-Net. The black box indicates the ground-truth motion region; the red box highlights where DAP-Net produces higher activations than baseline, capturing more discriminative motion cues.}
    \label{fig:visualization}
\end{figure}

\noindent  \textbf{Source distribution alignment metrics} measure the discrepancy between feature distributions of different sources. Centroid Distance quantifies the geometric distance between source feature centers. CORAL (Correlation Alignment) measures the discrepancy between source and target covariance matrices:
\begin{equation}
    \text{CORAL} = \frac{1}{4d^2} \|C_s - C_t\|_F^2,
\end{equation}
while Maximum Mean Discrepancy (MMD) evaluates the difference between source and target distributions in the reproducing kernel Hilbert space (RKHS). These metrics collectively provide insight into the model's ability to learn source-invariant representations.

All metrics are reported as relative percentage improvements. For increasing metrics (denoted as ↑), where higher values indicate better performance, such as Macro Accuracy, Micro Accuracy, Cross-Source Off-Diagonal Accuracy, and Cross-Source Accuracy, the improvement is computed as:
\begin{equation}
    \text{Improvement (\%)} = \frac{\text{Ours}-\text{Baseline}}{\text{Baseline}} \times 100\%.
\end{equation}
For decreasing metrics (denoted as ↓), where lower values indicate better alignment, such as Centroid Distance, CORAL, and MMD, the improvement is calculated as:
\begin{equation}
    \text{Improvement (\%)} = \frac{\text{Baseline}-\text{Ours}}{\text{Baseline}} \times 100\%.
\end{equation}
The results indicate that incorporating DAP leads to consistent improvements across all accuracy and cross-source metrics, with PointMLP achieving gains of 9–13\% and PST-Transformer 19–28\%. At the same time, cross-source distribution discrepancies are substantially reduced, with CORAL and MMD decreasing by up to 88\% and 90\%, respectively. Notably, Cross-Source Accuracy, which represents a more challenging evaluation under unseen heterogeneous sources, also shows significant improvement. \uline{These findings collectively demonstrate that DAP effectively mitigates inter-source distribution shifts, learns source-invariant action representations, and enhances the generalization capability of point cloud backbones in heterogeneous cross-source scenarios.}

\subsection{Visualization}
We have already presented extensive experiments in the main paper demonstrating the superiority of the proposed DAP-Net on the heterogeneous multi-source UniMM-HAR dataset. In addition, Fig.~\ref{fig:per_class_acc} visualizes the top 10 action categories with the largest accuracy improvements. Notably, actions with larger motion amplitudes show particularly pronounced gains, indicating that \uline{the cross-source consistency of Doppler patterns effectively enhances performance.}

As shown in Fig.~\ref{fig:visualization}, we compare the features extracted by DAP-Net with those obtained from the baseline (PointMLP). As expected, DAP-Net focuses more on regions with pronounced motion, which are more informative for action recognition, whereas the baseline exhibits weaker attention in these areas. These qualitative observations are consistent with the quantitative improvements, confirming that \uline{DAP-Net effectively leverages Doppler priors to concentrate on motion-relevant regions.}


\end{document}